\documentclass[10pt,journal,compsoc]{IEEEtran}
\usepackage{times}
\usepackage{soul}
\usepackage{url}
\usepackage[hidelinks]{hyperref}
\usepackage[utf8]{inputenc}
\usepackage[small]{caption}
\usepackage{graphicx}
\usepackage{amsmath}
\usepackage{amsthm}
\usepackage{booktabs}
\usepackage{color}
\usepackage{bm}
\usepackage{multirow}
\usepackage[table ]{ xcolor}
\usepackage{amsfonts,amssymb}
\usepackage{diagbox}
\usepackage{subcaption}
\usepackage[ruled,linesnumbered]{algorithm2e}
\usepackage{algpseudocode}
\usepackage{arydshln}
\usepackage{caption}
\usepackage{subcaption}
\usepackage[T1]{fontenc}
\newtheorem{thm}{Definition}[section]

\ifCLASSOPTIONcompsoc
  \usepackage[nocompress]{cite}
\else
  \usepackage{cite}
\fi
\ifCLASSINFOpdf
\else
\fi
\begin{document}
%
\title{Few-shot Partial Multi-view Learning}
%
%
%
%

\author{Yuan~Zhou,
        Yanrong~Guo,
        Shijie~Hao, 
        Richang~Hong,~\IEEEmembership{Senior Member,~IEEE,}
        Jiebo~Luo,~\IEEEmembership{Fellow,~IEEE}
\IEEEcompsocitemizethanks{\IEEEcompsocthanksitem Y. Zhou, Y. Guo, S. Hao, and R. Hong are with Key Laboratory of Knowledge Engineering with Big Data (Hefei University of Technology), Ministry of Education, and School of Computer Science and Information Engineering, Hefei University of Technology, Hefei 230009, China (e-mail: 2018110971@mail.hfut.edu.cn, yrguo@hfut.edu.cn, hfut.hsj@gmail.com, hongrc.hfut@gmail.com). J. Luo is with Department of Computer Science, University of Rochester, Rochester, NY 14627 (e-mail: jluo@cs.rochester.edu). Y. Guo is the corresponding author. \protect\\
}
\thanks{Manuscript received April 19, 2005; revised August 26, 2015.}}

%
%

\markboth{This paper has been accepted by IEEE Transactions on Pattern Analysis and Machine Intelligence.}%
{Shell \MakeLowercase{\textit{et al.}}: Bare Demo of IEEEtran.cls for Computer Society Journals}
%



\IEEEtitleabstractindextext{%
\begin{abstract}
It is often the case that data are with multiple views in real-world applications. Fully exploring the information of each view is significant for making data more representative. \textcolor{black}{However, due to various limitations and failures in data collection and pre-processing, it is inevitable for real data to suffer from view missing and data scarcity. The coexistence of these two issues makes it more challenging to achieve the pattern classification task.} Currently, to our best knowledge, few appropriate methods can well-handle these two issues simultaneously. Aiming to draw more attention from the community to this challenge, we propose a new task in this paper, called few-shot partial multi-view learning, which focuses on overcoming the negative impact of the view-missing issue in the low-data regime. The challenges of this task are twofold: (i) it is difficult to overcome the impact of data scarcity under the interference of missing views; (ii) the limited number of data exacerbates information scarcity, thus making it harder to address the view-missing issue in turn. \textcolor{black}{To address these challenges, we propose a new unified Gaussian dense-anchoring method. The unified dense anchors are learned for the limited partial multi-view data, thereby anchoring them into a unified dense representation space where the influence of data scarcity and view missing can be alleviated.} We conduct extensive experiments \textcolor{black}{to evaluate our method}. The results on Cub-googlenet-doc2vec, Handwritten, Caltech102, Scene15, Animal, ORL, tieredImagenet, and Birds-200-2011 datasets validate its effectiveness.
\end{abstract}

}

\maketitle

\IEEEdisplaynontitleabstractindextext

%
\IEEEpeerreviewmaketitle

\IEEEraisesectionheading{\section{Introduction}\label{sec:introduction}}
Data are often with multiple views in real-world applications as they can be observed from different viewpoints or captured by different sensors \cite{ding2014low,xu2015multi}. \textcolor{black}{Fully exploiting the information of each view is critical for building high-quality data representations}. However, real data inevitably suffer from \textcolor{black}{the view-missing issue} due to various failures in data preparation. \textcolor{black}{Moreover, it is time-consuming and expensive to obtain large-scale labeled data \cite{wang2020generalizing}. This makes it possible for the model to suffer from data scarcity in the target task as well.} These two issues affect each other and have a strong impact on the effectiveness of the pattern classification models.

\begin{figure}[t!]
\centering
\includegraphics[height=4.5cm]{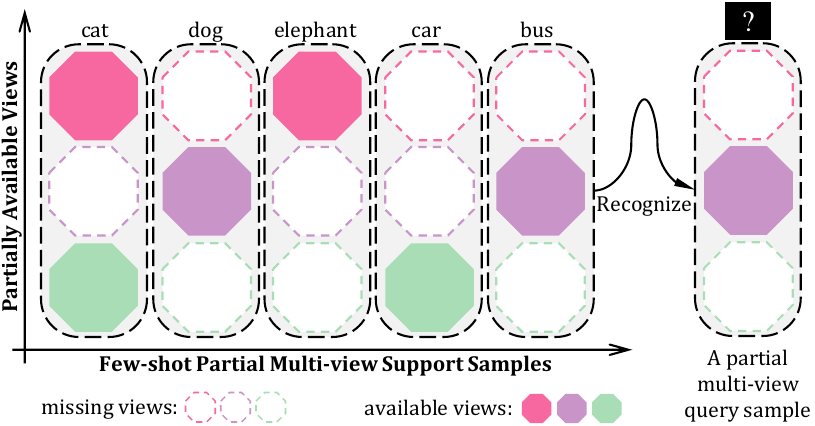}
\caption{An overview of our Few-shot Partial Multi-view Learning (FPML) task. The above figure is exemplified in the ``$5$-way $1$-shot $3$-view'' setting.}
\label{fig:1}
\end{figure}

In recent years, large efforts have been paid on partial multi-view learning \cite{li2014partial} with the goal of overcoming the influence of the view-missing issue as much as possible. The current methods for partial multi-view learning are mainly designed to address the partial multi-view data clustering \cite{li2014partial,zhao2016incomplete,zhang2018multi,hu2019doubly,wang2018partial,wang2020icmsc,xu2021adversarial} or the classification problem \cite{zhang2019cpm,zhang2020deep}. These methods generally assume that the samples are adequate.  \textcolor{black}{Thus, rich complementary information across samples can be explored, so as to help represent the incomplete multi-view data into a common low-dimensional subspace. Therefore, the view gaps of different samples can be reduced, and the negative effects caused by missing views can be overcome.} \par

Despite the success achieved by these methods, we find that they are unable to \textcolor{black}{handle the view-missing problem} in the low-data regime, \textcolor{black}{e.g., there only exists a labeled training sample in each category}. In this context, we propose a new task in this paper, \textcolor{black}{named} Few-shot Partial Multi-view Learning (FPML). \textcolor{black}{As described in Fig. \ref{fig:1}}, FPML aims to recognize novel unseen categories under the scenario that the data are scarce while being damaged by arbitrary view-missing. In other words, it aims to recognize new categories according to only a few labeled partial multi-view data. In the following, we first analyze why the current partial multi-view learning methods fail to address our proposed task. Then, we point out the challenges of FPML and give our solutions. \par

Specifically, the reasons for the failure of \textcolor{black}{the current partial multi-view learning methods in our task} can be concluded in two aspects. Firstly, it is impossible to directly map the incomplete multi-view data into a common subspace \textcolor{black}{under the low-data scenario}, as data scarcity makes it challenging to reduce the view gaps. For example, as shown in Fig.~\ref{fig:1}, \textcolor{black}{it is hard to reduce the view gaps between the samples ``cat'' and ``elephant'' as there is no any information about these two categories other than these two instances}. Secondly, the current methods aiming to address the view-missing issue are mainly based on the nonnegative matrix factorization \cite{li2014partial,zhao2016incomplete,zhang2018multi,hu2019doubly}, the generative adversarial network \cite{wang2018partial,xu2021adversarial}, and the auto-encoder \cite{wang2020icmsc,wang2021incomplete,wen2021cdimc,lin2021completer}. However, these methods all require sufficient data. The methods based on the generative adversarial network or the auto-encoder need sufficient data to optimize the learnable parameters of neural networks. In the low-data scenario, the methods based on the nonnegative matrix factorization are severely limited by the local optimum problem \cite{chao2017survey} as it is inaccessible to obtain accurate basis matrixes based on only one incomplete sample per category. \par

Overall, FPML faces two mutually-affected challenges. \textcolor{black}{On one hand}, it is difficult to alleviate the negative impact brought by data scarcity under \textcolor{black}{the view-missing problem}. \textcolor{black}{On the other hand}, the limited number of data raises the difficulty of overcoming \textcolor{black}{the view-missing issue} in turn. These two challenges affect each other, making them more challenging to address. To overcome these issues, we propose a new Unified Gaussian Dense-Anchoring (UGDA) method \textcolor{black}{by densely anchoring the limited partial multi-view data into a unified dense representation space.} \textcolor{black}{The first phase Dense Gaussian Anchor Inverse-aggregation (DGAI) assumes that the view observations satisfy the Gaussian distribution, and estimates the view distribution of the samples by retrieving their top-$k$ neighboring base categories in the available views. \textcolor{black}{The statistics of these retrieved neighboring categories are used to model the distribution of both the missing and the available views. Therefore, \textcolor{black}{the view-missing issue} can be relieved by the retrieved auxiliary statistics.}} \textcolor{black}{\textcolor{black}{Moreover}, the dense Gaussian anchors are sampled from the training sample distribution to densely anchor the training instances in each view, thereby alleviating the negative impact of data scarcity}. \textcolor{black}{\textcolor{black}{Besides, the inverse anchor aggregator is adopted to aggregate the Gaussian anchors or query embeddings of different views into a unified latent space. By leveraging the information in each view, the data can be more representative, which is beneficial for achieving robust classification results.}} Following DGAI, the Anchor Distribution Self-Rectification (ADSR) is conducted as the second phase, which aims to further adjust the representations of the Gaussian anchors and help to build a more effective anchor-based classifier. \par

\textcolor{black}{Aiming to demonstrate the effectiveness of our method, we conduct the extensive experiments on Cub-googlenet-doc2vec, Handwritten, Caltech102, Scene15, Animal, ORL, tieredImagenet, and Birds-200-2011. The results validate the effectiveness of our method. The contributions of this paper can be concluded as follows:}\par

\begin{itemize}
\item[$\circ$] \textcolor{black}{We propose a new task in this paper named Few-shot Partial Multi-view Learning (FPML), \textcolor{black}{aiming to facilitate} the research on overcoming the view-missing problem in the low-data regime.}
\item[$\circ$] \textcolor{black}{We provide an effective method for this task, namely Unified Gaussian Dense-Anchoring (UGDA).} \textcolor{black}{In UGDA, we propose to conduct the Dense Gaussian Anchor Inverse-aggregation (DGAI), thereby densely anchoring the \textcolor{black}{few-shot partial multi-view data} into a unified latent space \textcolor{black}{where the data-scarcity and the view-missing issue can be both relieved.}}
\item[$\circ$] \textcolor{black}{In addition, we also propose to conduct the Anchor Distribution Self-Rectification (ADSR). So, the latent anchors can be rectified to build a more effective anchor-based classifier.}
\item[$\circ$] \textcolor{black}{Last but not least, our UGDA has good compatibility. It can be combined with other FSL models and \textcolor{black}{make them more robust to the view-missing problem} in the low-data regime.}
\end{itemize}

\section{Related Work}
Partial Multi-view Learning (PML) and Few-Shot Learning (FSL) are two \textcolor{black}{typical relevant tasks to our} FPML. We first briefly review the methods of PML and FSL in Section 2.1 and Section 2.2. \textcolor{black}{Then, we discuss the relationships between the recent related works and our FPML in Section 2.3.} \par

\subsection{Partial Multi-view Learning}
\textbf{Methods based on nonnegative matrix factorization}. The first PML work was proposed by Li et al. \cite{li2014partial}, focusing on \textcolor{black}{addressing} the partial multi-view data clustering problem. \textcolor{black}{As the CCA-based methods \cite{hotelling1992relations,akaho2006kernel,andrew2013deep} are severely limited by the view-missing problem \cite{zhang2019cpm}}, Li et al. proposed to use the nonnegative matrix factorization \cite{lee1999learning} to learn a common subspace for the incomplete multi-view data. \textcolor{black}{Thus, the use of the complementarity across samples helps reduce the view gaps between the instances.} Following \cite{li2014partial}, many variants have been proposed. For example, \textcolor{black}{in order} to learn a better subspace, Zhao et al. \cite{zhao2016incomplete} designed a probabilistic Laplacian term to improve the subspace's compactness. Xu et al. \cite{xu2018partial} proposed a reconstruction term, \textcolor{black}{thereby guaranteeing the effectiveness of the subspace features.} Zhang et al. \cite{zhang2018multi} proposed a block-diagonal regularizer \textcolor{black}{that makes the subspace representations more discriminative.} Different from these methods, Hu et al. \cite{hu2019doubly} \textcolor{black}{advanced} \cite{li2014partial} by introducing the semi-nonnegative matrix factorization, \textcolor{black}{thus making it possible for models to process negative entries.} \par

\textbf{Methods based on \textcolor{black}{generative adversarial network} or auto-encoder}. In addition to the nonnegative matrix factorization, there are also some methods based on the generative adversarial network \cite{wang2018partial,xu2021adversarial} or the auto-encoder \cite{wang2020icmsc,wang2021incomplete,wen2021cdimc,lin2021completer}. \cite{wang2018partial,xu2021adversarial} proposed to train a generator to impute the missing views before mapping them into the subspace. In \cite{wang2020icmsc,wang2021incomplete,wen2021cdimc,lin2021completer}, an auto-encoder was trained to bridge the gaps between different views, while a decoder equipped with the reconstruction loss was employed to ensure the effectiveness of the feature encoding. \textcolor{black}{Specifically}, Wen et al. \cite{wen2021cdimc} proposed to build \textcolor{black}{a deeper} encoder \textcolor{black}{so as to make the} subspace features more \textcolor{black}{discriminative}. Wang et al. \cite{wang2020icmsc} designed the exclusive self-expression module, \textcolor{black}{thus improving the representations of the subspace}. Lin et al. \cite{lin2021completer} enhanced the subspace via maximizing the lower bound of the mutual information between different views. Wang et al. \cite{wang2021incomplete} used the graph networks to model the \textcolor{black}{relationships across views so as to make data} more representative. Aiming to solve the partial multi-view data classification problem, Zhang et al. \cite{zhang2019cpm} recently proposed the Cross Partial Multi-view (CPM) model, which achieves a better trade-off between the views' consistency and complementarity via mimicking the data transmission procedure. Also, by additionally adopting the adversarial strategy, Zhang et al. \cite{zhang2019cpm} further introduced CPM-GAN \cite{zhang2020deep}. Like \cite{wang2020icmsc,wang2021incomplete,wen2021cdimc,lin2021completer}, the reconstruction loss is also utilized in \cite{zhang2019cpm,zhang2020deep} to ensure the quality of the subspace representations.

\subsection{Few-shot Learning}
\textbf{Methods based on data augmentation}. FSL has attracted \textcolor{black}{much attention from the community in the past few years}, which aims to help machine learning models quickly learn novel visual concepts from only a few labeled training data. \textcolor{black}{Thereby, the expenses spent on data collection and annotation can be reduced}. In order to solve this task, researchers proposed to directly augment training samples using the data augmentation strategies. For example, \cite{wang2018low,zhang2019few,li2020adversarial} proposed to synthesize new data from \textcolor{black}{the limited labeled samples}. \cite{wu2018exploit,douze2018low} proposed to retrieve the auxiliary data from the relevant datasets. Despite the straightforwardness and intuitiveness of \textcolor{black}{these methods, the limitation is that} the augmentation policies often should be specifically designed for each dataset due to the domain gaps \cite{wang2020generalizing}. \par

\textbf{\textcolor{black}{Methods based on metric learning}}. Due to the high effectiveness and flexibility, \textcolor{black}{the metric-learning-based methods have dominated in FSL}, which aim at building a metric-based classifier from limited \textcolor{black}{data. \cite{vinyals2016matching,snell2017prototypical} are two typical  works. Specifically}, \cite{vinyals2016matching} built a matching-based classifier that matches the test sample to \textcolor{black}{the training instance most likely sharing the same label;} \cite{snell2017prototypical} proposed a prototypical classifier using the prototypes to approximate the representations of categories. Based on \textcolor{black}{these two works, many variant methods have been proposed}. For example, \cite{li2019few,wang2020cooperative} proposed to advance the metric-based classification by explicitly leveraging the class-wise relationships. \cite{li2020boosting,wang2020large} resorted to Glove \cite{pennington2014glove} and word2vec \cite{mikolov2013efficient} \textcolor{black}{to fully explore the semantic information of samples.} Pahde et al. \cite{pahde2021multimodal} proposed to additionally consider textual information via using the adversarial strategy \textcolor{black}{which makes the} prototypes more representative. Zhang et al. \cite{zhang2021prototype} used primitive knowledge to \textcolor{black}{enhance the prototypical} representations. Tian et al. \cite{tian2020rethinking} proposed to train embedding networks using the traditional supervised pre-training, \textcolor{black}{which leads to  better feature embeddings}. Finn et al. \cite{finn2017model} proposed \textcolor{black}{a model-agnostic meta-learning approach that uses} the second-order-based optimization to realize the fast adaptation of models to unseen classes. Recently, Nuthalapati et al. \cite{nuthalapati2021multi} successfully applied the FSL technique \textcolor{black}{to  agriculture and realized effective pest detection.} \par

\textcolor{black}{\textbf{Methods based on external memory}. Inspired by Neural Turing Machine \cite{graves2014neural},  \textcolor{black}{Santoro et al. \cite{santoro2016meta} proposed a memory-based FSL approach, called memory-augmented neural networks, which categorizes the queries by continuously updating its key-to-value memory module. \cite{kaiser2017learning, ramalho2019adaptive} proposed to reduce the redundancy of the memory, thereby further improving categorization accuracy.} Different from \cite{santoro2016meta,kaiser2017learning, ramalho2019adaptive}, Cai et al. \cite{cai2018memory} used the memory to generate the task-specific convolution kernels, so that the queries can be \textcolor{black}{encoded} in a more effective way. \textcolor{black}{Zhu et al. \cite{zhu2018compound} proposed to store the memory in the matrix format, and thus the memory retrieval's efficiency can be improved.}}

\subsection{\textcolor{black}{Discussion}}
\textbf{\textcolor{black}{Limitations of current partial multi-view learning and few-shot learning methods}}. \textcolor{black}{The current Partial Multi-view Learning (PML) or Few-Shot Learning (FSL) methods cannot well-handle our proposed task. On one hand, the view-missing issue is ignored in FSL. The FSL methods generally assume that the data are in a single view and the data observations are all complete. Thus, they are not ready for overcoming the negative impact of missing views. On the other hand, although the PML methods focus on \textcolor{black}{addressing the view-missing issue}, they are not designed to tackle the low-data task, e.g., there are only one or few labeled instances per category. Data scarcity will severely limit their effectiveness in solving the view-missing problem.} \par

\textcolor{black}{\textbf{Relationships between partial multi-view learning, few-shot learning, and our few-shot partial multi-view learning}}. \textcolor{black}{PML and FSL aim to address the view-missing and the data-scarcity issue respectively. However, it is inevitable for real data to suffer from these two problems simultaneously in real-world applications}. \textcolor{black}{To tackle this challenge}, we propose the Few-shot Partial Multi-view Learning (FPML) task in this paper, through considering the coexistence and interaction of data scarcity and view missing. \textcolor{black}{For the current research in PML and FSL, there also exist different research preferences}. In FSL, some methods \textcolor{black}{target on} training a better feature extractor, \textcolor{black}{thus leading to producing high-quality feature representations}, such as \cite{rizve2021exploring,Mazumder_2021_WACV}. However, most of the current PML works do not aim at building a better feature extractor or view sensor. Instead, the view observations are generally extracted by \textcolor{black}{the common approaches}, such as LBP, HOG, and Gabor. \textcolor{black}{Following these PML methods, we do not rely on \textcolor{black}{employing} better feature exactors in our work}.

\textcolor{black}{In addition to FSL, there are some works that aim at extending the FSL problem to a more realistic scenario, such as \cite{jiang2020few,tsimpoukelli2021multimodal,zhao2021few,tao2020few}. \cite{zhao2021few} proposed to additionally consider the label-ambiguity problem in FSL. \cite{jiang2020few,tsimpoukelli2021multimodal} extended FSL to the multi-view or multi-modal scenario. \cite{tao2020few} endued the FSL models with the capability of incremental learning via mimicking the human-like learning manner. Our work is different from these related ones. On one hand, compared with \cite{jiang2020few,tsimpoukelli2021multimodal}, our FPML additionally considers the influence coming from the view-missing issue. On the other hand, rather than focusing on addressing the label-ambiguity problem or extending FSL to the incremental learning scenario in \cite{zhao2021few,tao2020few}, we study on how to alleviate the view-missing issue in the low-data regime.}

\begin{figure*}[t!]
    \centering
    \includegraphics[height=5.2cm]{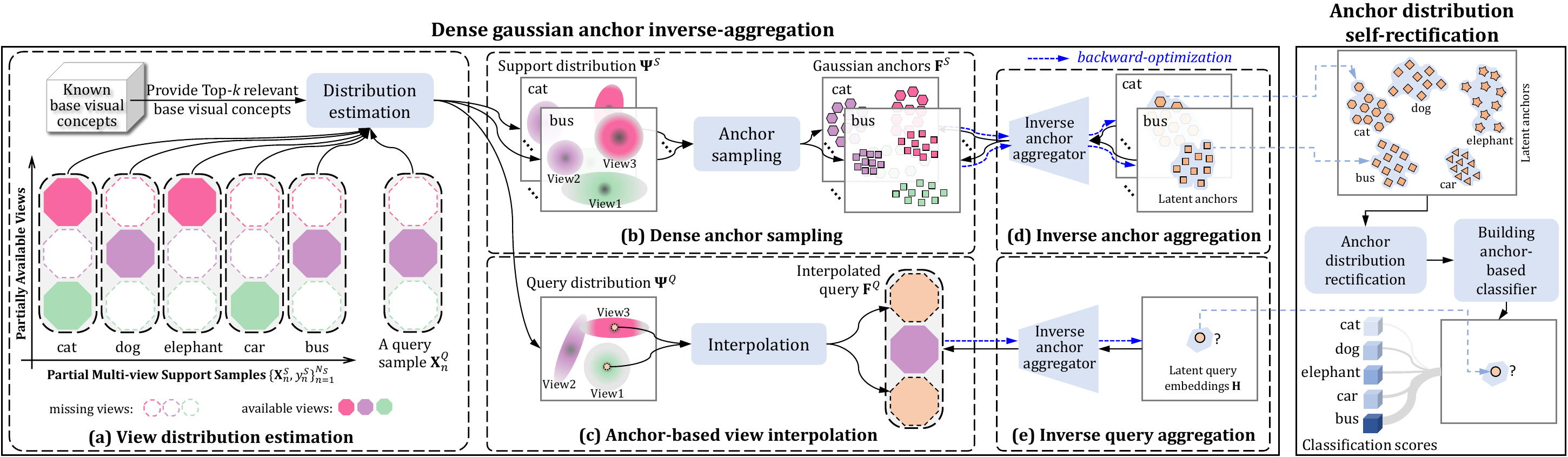}
    \caption{\textcolor{black}{The illustration of our unified Gaussian dense-anchoring approach. The dense gaussian anchor inverse-aggregation aims to learn the dense latent anchor representations for the low-shot supports and build the latent query representations. The anchor distribution self-rectification further calibrates the latent anchors and constructs an anchor-based classifier to infer the label of the query. The figure is best viewed in color.}}
    \label{fig:2}
    \end{figure*}

\section{\textcolor{black}{Few-shot Partial Multi-view Learning}}
\textcolor{black}{We elaborate on our FPML task in this section. First, we introduce the preliminaries in Section 3.1. Then, the definition of the task is given in Section 3.2. Finally, we introduce the details about the base visual concepts used in knowledge transfer. Notice that the main notations and the corresponding definitions are summarized in Table. \ref{tab:0}.} \par

\subsection{Preliminaries}
\textcolor{black}{\textbf{Partial multi-view supports and queries}. An FPML task contains a support set $\bm{\mathrm{S}}$ and a query set $\bm{\mathrm{Q}}$. The support set $\bm{\mathrm{S}}$ consists of a few labeled multi-view samples, $\bm{\mathrm{S}}=\big\{\bm{\mathrm{X}}^{S}_{n},\mathit{y}^{S}_{n}\big\}_{n=1}^{N_{S}}$ where $\bm{\mathrm{X}}_{n}^{S}$ denotes the observations about the $n$-$th$ support sample and $y_{n}^{S}$ indicates the label of $\bm{\mathrm{X}}^{S}_{n}$. The query set $\bm{\mathrm{Q}}$ is composed of the samples whose labels are unknown, denoted as $\bm{\mathrm{Q}}=~\big\{\bm{\mathrm{X}}^{Q}_{n}\big\}_{n=1}^{N_{Q}}$. In our task, the support sample and the query sample may be both partially view-missing. Therefore, they can be represented as follows:  $\bm{\mathrm{X}}^{S}_{n}=\big\{\bm{x}^{S}_{n,v}*\omega^{S}_{n,v}\big\}_{v=1}^{V}$ and $\bm{\mathrm{X}}^{Q}_{n}=\big\{\bm{x}^{Q}_{n,v}*\omega^{Q}_{n,v}\big\}_{v=1}^{V}$. $\bm{x}^{S}_{n,v}$ and $\bm{x}^{Q}_{n,v}$ indicate the $v$-$th$ view of the samples $\bm{\mathrm{X}}^{S}_{n}$ and $\bm{\mathrm{X}}^{Q}_{n}$, while $\omega^{S}_{n,v}$ and $\omega^{Q}_{n,v}$ represent the corresponding view indicators. Specially, if the view $\bm{x}^{S}_{n,v}$ or $\bm{x}^{Q}_{n,v}$ is missing, $\omega^{S}_{n,v}=0$ or $\omega^{Q}_{n,v}=0$; otherwise, $\omega^{S}_{n,v}=1$ or $\omega^{Q}_{n,v}=1$. Following the paradigm of FSL \cite{snell2017prototypical,vinyals2016matching}, we consider the well-organized support set $\bm{\mathrm{S}}$ in our FPML, i.e., the ``$|\bm{\mathrm{C}}|$-way $K$-shot $V$-view'' setting. In other words, $\bm{\mathrm{S}}$ is prepared by sampling $K$ samples from each of the categories $\bm{\mathrm{C}}$ under the $V$ views. Here, the $V$ views might be partially available for each sample.} \par

\textbf{\textcolor{black}{View-missing rate}}. \textcolor{black}{To quantify the degree of view missing, the view-missing rate $\eta$ is employed in our FPML.} It is defined as the proportion between the missing views and all the missing and \textcolor{black}{the} available ones in an episode,
\begin{align}
\eta=1-\frac{\sum^{V}_{v=1}(\sum^{N_{S}}_{n=1}\omega^{S}_{n,v}+\sum^{N_{Q}}_{n=1}\omega^{Q}_{n,v})}{(N_{Q}+N_{S})\times V}.
\label{eq:2}
\end{align}
\textcolor{black}{In the equation, $\omega^{S}_{n,v}$ and $\omega^{Q}_{n,v}$ represent the view indicators, and  $N_S$ and $N_Q$ indicate the number of the supports and the queries.}

\subsection{\textcolor{black}{Task Definition}}
\textcolor{black}{In this subsection, we first give the definition of FPML, and then provide a discussion for this definition.}
\begin{thm}[\textit{\textbf{Few-shot Partial Multi-view Learning}}]
\textcolor{black}{FPML aims to recognize an unseen multi-view query sample $\bm{\mathrm{X}}^{Q}_{n}$ according to only a few labeled multi-view supports in the support set $\bm{\mathrm{S}}=\big\{\bm{\mathrm{X}}^{S}_{n}, y_n^S\big\}_{n=1}^{N_S}$, under the scenario that the views of $\bm{\mathrm{X}}^{Q}_{n}$ and $\bm{\mathrm{X}}^{S}_{n}$ can be both partially missing.}
\end{thm}

\textcolor{black}{Actually, the above task definition can also be mathematically described by Eq. \ref{eq:1},}
\begin{align}
\textcolor{black}{y^{Q}_{n}=\mathop{\arg\max}_{y'\in\bm{C}} P(y'|\bm{\mathrm{X}}^{Q}_{n},\bm{\mathrm{S}})}
\label{eq:1}
\end{align}
\textcolor{black}{where $P(y'|\bm{\mathrm{X}}^{Q}_{n},\bm{\mathrm{S}})$ indicates the probability that the query sample $\bm{\mathrm{X}}^{Q}_{n}$ is classified as the category $y'$ according to the information in the support set $\bm{\mathrm{S}}$. \textcolor{black}{From the above definition, it is easy to draw the following observations. On one hand, as the support set is very small, there is only limited categorization information that can be used in classifying the queries. On the other hand, the partially available views severely damage the consistency of the data, which poses more difficulties to the data-driven machine-learning models.}}

\begin{table}
    \centering
    \caption{The main notations and \textcolor{black}{the corresponding} definitions.}
    \setlength{\tabcolsep}{0.5mm}{
    \begin{tabular}{l|l}
        \toprule
        Notation & \multicolumn{1}{c}{Definition} \\
        \midrule
       $\bm{\mathrm{S}}/\bm{\mathrm{Q}}$ & The support$/$query set. \\
       $\bm{\mathrm{X}}^{S}_{n}/\bm{\mathrm{X}}^{Q}_{n}$ & The $n$-$th$ support$/$query sample. \\
       $\bm{x}^{S}_{n,v}/\bm{x}^{Q}_{n,v}$  & The observations about the $v$-$th$ view of $\bm{\mathrm{X}}^{S}_{n}/\bm{\mathrm{X}}^{Q}_{n}$. \\
       $\omega^{S}_{n,v}/\omega^{Q}_{n,v}$ & The view indicator about $\bm{x}^{S}_{n,v}/\bm{x}^{Q}_{n,v}$. \\
       $\bm{\Psi}^{S}/\bm{\Psi}^{Q}$ & The view distribution of the supports$/$queries. \\
       $y^{S}_{n}/y^{Q}_{n}$ & The label$/$prediction of $\bm{\mathrm{X}}^{S}_{n}/\bm{\mathrm{X}}^{Q}_{n}$. \\
       $\bm{\mathrm{F}}^{S}$ & The Gaussian anchors generated in an episode. \\
       $\bm{\mathrm{F}}^{Q}$ & The queries after being interpolated. \\
       $\bm{\mathrm{A}}$ & The anchors that are aggregated into the latent space. \\
       $\bm{\mathrm{H}}$ & The latent embeddings of the query samples. \\
       $\bm{\mathrm{C}}_{b}$ & The base visual concepts. \\
       $\ddot{\bm{\mu}}_{c,v}/\ddot{\bm{\varsigma}}_{c,v}$ & The mean$/$covariance of the base class $c$ in the  $v$-$th$ view. \\
       $\bm{\mathrm{C}}$ & The categories of the target FPML task.  \\
       $\bm{\vartheta}_{c}$ & The distribution offset for the anchors of the class $c$. \\
       $\mathcal{L}_{ce}$ & The cross-entropy loss. \\
       $\mathcal{L}_{se}$ & The Shannon-entropy loss. \\
       $\mathcal{L}$ & The aggregation loss. \\
       $\bm{\delta}_{c}$ & The mean latent anchor representations of the class $c$. \\
       $\zeta_{c}$ & The global query relation score about the class $c$. \\
       $\bm{\alpha}_m/\bm{\beta}_m$ & The relation scores of the $m$-$th$ latent anchor$/$query to $\bm{\mathrm{C}}$. \\
       $\bm{\mathrm{W}}/\bm{\mathrm{B}}$ & The learnable weights$/$biases of the aggregation evaluator. \\
        \bottomrule
    \end{tabular}}
    \label{tab:0}
\end{table}

\subsection{Base Visual Concepts for Knowledge \textcolor{black}{Transfer}}
\textbf{Base visual concepts}. \textcolor{black}{Following the paradigm of FSL, the base visual concepts $\bm{\mathrm{C}}_{b}$ are used in our FPML to make it possible for knowledge  transfer. For brevity, we term the data set about these base visual concepts $\bm{\mathrm{C}}_{b}$ as the base set $\bm{\mathrm{D}}_{b}$. The categories of the base set $\bm{\mathrm{D}}_{b}$ are disjoint to these of the target FPML task, i.e., $\bm{\mathrm{C}}_{b}\cap\bm{\mathrm{C}}=\emptyset$. We follow FSL works, such as \cite{hu2021leveraging,boudiaf2020information,liu2020prototype}, and assume there are the sufficient auxiliary base samples that can be obtained.} \par

\textcolor{black}{\textbf{Transferable statistical information}}. \textcolor{black}{By assuming that the view observations satisfy the Gaussian distribution, we explicitly extract the statistics of the base visual concepts $\bm{\mathrm{C}}_{b}$ to facilitate the subsequent process of building the dense Gaussian anchors. We calculate the mean and the covariance of the base categories in each view through Eq. \ref{eq:3} and Eq. \ref{eq:4},}
\begin{align}
\ddot{\bm{\mu}}_{c,v} = \frac{1}{|\bm{\mathrm{O}}_{c,v}|}\sum_{\bm{x}\in\bm{\mathrm{O}}_{c,v}} \bm{x}
\label{eq:3}
\end{align}
\begin{align}
\ddot{\bm{\varsigma}}_{c,v}= \frac{1}{|\bm{\mathrm{O}}_{c,v}|-1}\sum_{\bm{x}\in\bm{\mathrm{O}}_{c,v}}(\bm{x}-\ddot{\bm{\mu}}_{c,v})(\bm{x}-\ddot{\bm{\mu}}_{c,v})^{\top}
\label{eq:4}
\end{align}
where $\bm{\mathrm{O}}_{c,v}$ denotes the sample set of the base class $c$ in the $v$-$th$ view, and $\ddot{\bm{\mu}}_{c,v}\in \mathbb{R}^{d_v\times 1}$ and $\ddot{\bm{\varsigma}}_{c,v}\in\mathbb{R}^{d_v\times d_v}$ denote the mean and the covariance about the $v$-$th$ view observations of the base class $c$. \textcolor{black}{Based on the above equations, the statistics of all the base categories can be obtained $\left\{\big\{\ddot{\bm{\mu}}_{c,v}, \ddot{\bm{\varsigma}}_{c,v}\big\}_{c=1}^{|\bm{\mathrm{C}}_{b}|}\right\}_{v=1}^{V}$. This explicit statistical information can help transfer the prior knowledge of the auxiliary base visual concepts to the target task more effectively, which is beneficial for building the high-quality anchors and boosting the model's robustness to view missing and data scarcity.}

\section{Methodology}
\textcolor{black}{In this section, we elaborate on our proposed UGDA method. First, we provide an overview for our approach in Section 4.1. Then, in Section 4.2 and Section 4.3, we introduce the details of the key components. }

\subsection{Method Overview}
Our UGDA consists of the two phases, which are the Gaussian Anchor Inverse-Aggregation (DGAI) and the Anchor Distribution Self-Rectification (ADSR). \textcolor{black}{The workflow of our method is shown in Fig. \ref{fig:2} and the pseudo codes are provided in Algorithm 1 at the end of this section.}

\textcolor{black}{In the DGAI phase, we first estimate the view distribution of the supports and the queries by retrieving their top-$k$ neighboring base categories in the available views. The statistics of these neighboring categories can help model the distribution of the available views. Moreover, they are also beneficial for approximating the distribution of the missing views due to the commonality between the missing and the available ones, i.e., the sample's missing views should be consistent with its available ones and have relatively high correlations to these retrieved neighboring categories.} \textcolor{black}{Therefore, the influence of the view-missing issue can be relieved.} \textcolor{black}{Afterwards, the Gaussian anchors are densely sampled from the distribution of the supports, so as to overcome the influence of data scarcity. Furthermore}, the missing views of the queries are interpolated by the anchors at the center of the corresponding view distribution. \textcolor{black}{Finally, the inverse anchor aggregator is employed to leverage the view-specific and the view-common information in an implicit way, which helps represent the anchors or query embeddings of different views into a unified latent space and reduces the influence of the view-missing issue.} In this way, \textcolor{black}{the unified data representations are learned,} and \textcolor{black}{the limited incomplete supports} are densely anchored in the latent space. The ADSR phase further rectifies the distribution of the latent anchors by jointly using \textcolor{black}{the supervised cross-entropy term and the unsupervised Shannon-entropy term. The rectified anchors are utilized to build the anchor-based classifier to infer the categories of the queries.} \par

\subsection{Dense Gaussian Anchor Inverse-aggregation}
\textcolor{black}{The DGAI phase contains the five stages, i.e.,} the view distribution estimation, the dense anchor sampling, the anchor-based view interpolation, the inverse anchor aggregation, and the inverse query aggregation. \textcolor{black}{In the following subsections, we give the details about these stages.}

\subsubsection{View Distribution Estimation}
\textcolor{black}{In the first stage, we estimate the view distribution of the supports and the queries by fully considering their available view information. For brevity,} we take the sample $\bm{\mathrm{X}}=\big\{\bm{x}_{v}*\omega_{v} \big\}_{v=1}^{V}\in \big\{\bm{\mathrm{S}}\cup\bm{\mathrm{Q}}\big\}$ as an example to illustrate this process. Of note, if $\bm{\mathrm{X}}$ is a query sample, its category is unknown; otherwise, the category of $\bm{\mathrm{X}}$ is denoted as $y$.

Specifically, we first measure the distance of the sample $\bm{\mathrm{X}}$ against all the base visual concepts $\bm{\mathrm{C}}_{b}$ in its available views using Eq. \ref{eq:5},
\begin{equation}
\ddot{\sigma}_{c,v}=
\begin{cases}
\mathrm{Euclidean}(\bm{x}_{v}, \ddot{\bm{\mu}}_{c,v}) & \mbox{if} \,\,\, \omega_{v}\neq 0 \\
\# \mathrm{Pass} & \mbox{else}.
\end{cases}
\label{eq:5}
\end{equation}
\textcolor{black}{In the above equation,} $\mathrm{Euclidean}(\cdot, \cdot)$ denotes the Euclidean distance function, and ``$\#\mathrm{Pass}$'' indicates ignoring the distance calculation procedure due to view missing. Also, the scalar $\ddot{\sigma}_{c,v}$ denotes the distance between the sample $\bm{\mathrm{X}}$ and the base class $c$ in the $v$-$th$ view. \textcolor{black}{Here, $\ddot{\sigma}_{c,v}\in\ddot{\bm{\sigma}}_{v}$ where $\ddot{\bm{\sigma}}_{v}=\big\{\ddot{\sigma}_{c,v}\big\}_{c=1}^{|\bm{\mathrm{C}}_{\mathrm{b}}|}$ is the set of the distance information obtained from the $v$-$th$ view}. \textcolor{black}{Specially, we define $\ddot{\bm{\sigma}}_{v}=\emptyset$ if the $v$-$th$ view observations of $\bm{\mathrm{X}}$ are missing. Therefore, the distance of the sample $\bm{\mathrm{X}}$ against the base categories $\bm{\mathrm{C}}_{b}$ in its available views can be represented as~$\big\{\ddot{\bm{\sigma}}_{v}\big\}_{v=1}^{V}$.}

\textcolor{black}{After measuring the distance between the sample $\bm{\mathrm{X}}$ and the base visual concepts $\bm{\mathrm{C}}_{b}$}, we retrieve the categories that are top-$k$ \textcolor{black}{neighboring} to $\bm{\mathrm{X}}$ in the available views, \textcolor{black}{which can be described by Eq. \ref{eq:6a} and Eq. \ref{eq:6b}:}
\begin{equation}
\textcolor{black}{\bm{\mathrm{J}}=\bigcup_{v=1}^{V}\bm{\Delta}_v},
\label{eq:6a}
\end{equation}
\begin{equation}
\textcolor{black}{\bm{\Delta}_v=}
\begin{cases}
\textcolor{black}{\mathrm{Retrieve}(\ddot{\bm{\sigma}}_{v}|k)} & \textcolor{black}{\mbox{if} \,\,\, \omega_{v}\neq 0} \\
\textcolor{black}{\emptyset} & \textcolor{black}{\mbox{else}}.
\end{cases}
\label{eq:6b}
\end{equation}
\textcolor{black}{In the above equations, $\bm{\mathrm{J}}=\big\{J_{n}\big\}_{n=1}^{N_{\mathrm{J}}}$ indicates the indexes of the selected base categories for the sample $\bm{\mathrm{X}}$. The operator $\mathrm{Retrieve}(\cdot|k)$ means selecting the top-$k$ relevant base visual concepts to $\bm{\mathrm{X}}$ in the available view and returning the corresponding indexes $\bm{\Delta}_v$. Of note, if the observations of the $v$-$th$ view of $\bm{\mathrm{X}}$ are missing, $\bm{\Delta}_v=\emptyset$.}

We use $\big\{\Psi_{v}\big\}_{v=1}^{V}$ to denote \textcolor{black}{the view distribution of the sample} $\bm{\mathrm{X}}$. We assume the view observations obey the multivariate Gaussian distribution, and thus $\Psi_{v}$ satisfies the condition $\Psi_{v}\sim \mathcal{N}(\bm{\mu}_{v},\bm{\varsigma}_{v})$. \textcolor{black}{Accordingly, $\Psi_{v}$ can be estimated by calculating the values of the} mean $\bm{\mu}_{v}$ and the covariance $\bm{\varsigma}_{v}$ \textcolor{black}{under the guidance of the statistics} from the retrieved neighboring categories $\big\{ \ddot{\bm{\mu}}_{J, v},\ddot{\bm{\varsigma}}_{J, v}\big\}_{J\in \bm{\mathrm{J}}}$,
\begin{equation}
\bm{\mu}_{v}=
\begin{cases}
\frac{1}{N_{\mathrm{J}}+1}\left[\bm{x}_{v}+\sum\limits_{J\in\bm{\mathrm{J}}} \ddot{\bm{\mu}}_{J,v} \right] & \mbox{if} \,\,\, \omega_{v}\neq 0 \\
\frac{1}{N_{\mathrm{J}}}\sum\limits_{J\in\bm{\mathrm{J}}} \ddot{\bm{\mu}}_{J,v} & \mbox{else},
\end{cases}
\label{eq:8}
\end{equation}
\begin{equation}
\bm{\varsigma}_{v}=\frac{1}{N_{\mathrm{J}}}\sum_{J\in\bm{\mathrm{J}}}\ddot{\bm{\varsigma}}_{J,v}.
\label{eq:9}
\end{equation}
\textcolor{black}{By repeatedly applying the above process to all the samples, the view distribution of the supports and the queries can be generated, represented as $\bm{\Psi}^{S}=~\left\{\big\{\Psi_{n,v}^{S}\big\}_{v=1}^{V}, y_{n}^{S}\right\}_{n=1}^{N_{S}}$ and $\bm{\Psi}^{Q}=~\left\{\big\{\Psi_{n,v}^{Q}\big\}_{v=1}^{V}\right\}_{n=1}^{N_{Q}}$ respectively. $\Psi_{n,v}^{S}$ or $\Psi_{n,v}^{Q}$ denotes the $v$-$th$ view distribution of the $n$-$th$ support or query. In addition}, $\Psi_{n,1}^{S}$...$\Psi_{n,V}^{S}$ have a common label $y_{n}^{S}$ as all of them are the distribution of the sample $\bm{\mathrm{X}}_{n}^{S}$. The category of the distribution $\Psi_{n,v}^{Q}$ is unknown because the labels of the queries are inaccessible. \par

\subsubsection{Dense Anchor Sampling}
\textcolor{black}{After the first stage, the Gaussian anchors are densely sampled from the support sample distribution}, thereby densely anchoring the training instances individually in each view. The details about sampling the Gaussian anchors can be described by Eq. \ref{eq:10},
\begin{equation}
\bm{\mathrm{F}}^{S}, \bm{\mathrm{Y}}^{S}= \bigcup_{v=1, n=1}^{V,N_S} \mathrm{Anchor\_sampler}(\Psi^{S}_{n,v},y_{n}^{S}|N_{F}).
\label{eq:10}
\end{equation}
Specifically, in the equation, $\mathrm{Anchor\_sampler}(\Psi^{S}_{n,v},y_{n}^{S}|N_{F})$ indicates sampling the $N_{F}$ anchors ($s.t.\,\,\, N_{F}\gg 1$) from the \textcolor{black}{Gaussian} distribution \textcolor{black}{$\Psi^{S}_{n,v}$. $\bm{\mathrm{F}}^{S}=~\big\{\bm{\mathrm{F}}^{S}_{v}\big\}_{v=1}^{V}$ denotes the set of the anchors sampled from all the support samples, where $\bm{\mathrm{F}}_{v}^{S}=~\big\{\bm{\mathit{f}}_{m,v}^{S}\big\}_{m=1}^{N_S\times N_F}$ is the \textcolor{black}{anchor set about} the $v$-$th$ view. Additionally}, $\bm{\mathrm{Y}}^{S}$ represents the category labels of these generated anchors. \textcolor{black}{We define the label of the anchor as the category of the distribution that it belongs to. More specifically}, if an anchor is sampled from the distribution $\Psi_{n,v}^{S}$, it has the same label as the distribution $\Psi_{n,v}^{S}$. \textcolor{black}{Specially, the label of $\Psi_{n,v}^{S}$ is irrelevant to the subscript $v$, as $\Psi_{n,1}^{S}$...$\Psi_{n,V}^{S}$ have a common label $y_{n}^{S}$, i.e., the label of the $n$-$th$ support. Therefore, $\bm{\mathrm{Y}}^{S}$ can be re-written as $\bm{\mathrm{Y}}^{S}=~\big\{\bm{\mathrm{Y}}^{S}_{v}\big\}_{v=1}^{V}$ where $\bm{\mathrm{Y}}^{S}_{1}=...=\bm{\mathrm{Y}}^{S}_{V}=\big\{y^{S}_{m}\big\}_{m=1}^{N_S\times N_F}$.} \textcolor{black}{After the dense sampling of the Gaussian anchors from the support sample distribution, the original $N_S$ supports can be augmented by the $N_S\times N_F$ anchor instances (notice that $N_S\times N_F>>N_S$). Thus, these limited supports can be anchored into the representation space in a dense way. By jointly considering the information of these anchors, the data-scarcity problem of the support set can be alleviated.}
\par

\subsubsection{Anchor-based View Interpolation}
\textcolor{black}{At the same time as sampling the dense anchors, the anchor-based view interpolation is conducted to complete the missing views of the queries by considering the central anchor representations of the estimated view distribution. This process can be described by the following equation,}
\begin{equation}
\bm{\mathit{f}}_{n,v}^{Q}=
\begin{cases}
\bm{x}^{Q}_{n,v} & \mbox{if} \,\,\, \omega^{Q}_{n,v}\neq 0 \\
\mathrm{Center\_sampler}(\Psi^{Q}_{n,v}) & \mbox{else}.
\end{cases}
\label{eq:11}
\end{equation}
In this equation, $\mathrm{Center\_sampler}(\Psi^{Q}_{n,v})$ represents sampling the anchor located at the center of the distribution $\Psi^{Q}_{n,v}$. For simplicity, the updated queries are denoted as $\bm{\mathrm{F}}^Q=\big\{\bm{\mathrm{F}}^{Q}_{v}\big\}_{v=1}^{V}$ where $\bm{\mathrm{F}}_{v}^{Q}=\big\{\bm{\mathit{f}}_{n,v}^{Q}\big\}_{n=1}^{N_{Q}}$ indicates the $N_{Q}$ query samples observed in the $v$-$th$ view. \par

\subsubsection{\textcolor{black}{Inverse Anchor Aggregation}}
After the above stages, the anchors of different views are aggregated into a unified latent space, so as to further reduce the bias induced by missing views. To realize the effective anchor aggregation, \textcolor{black}{we first introduce the aggregation loss, and then propose the Inverse Anchor Aggregator (IAA) based on the backward-aggregation manner \cite{park2019deepsdf,tan1995reducing}. In the following, we first introduce the aggregation loss and IAA, and then elaborate on the workflow of our inverse anchor aggregation.}

\vspace{0.3cm}
\noindent\textcolor{black}{$(a).$ \textit{Aggregation Loss}}
\vspace{0.15cm}

\textcolor{black}{In this part, we first give the formal definition of the aggregation loss in Definition 4.1. Then, we provide a discussion for this definition.}
\begin{thm}[\textit{\textbf{Aggregation Loss}}]
\textcolor{black}{Assume that a sample $\bm{\mathrm{X}}$ is composed of multi-view features, i.e., $\bm{\mathrm{X}}=\big\{\bm{\mathit{f}}_{v}\big\}_{v=1}^{V}$}. When aggregating \textcolor{black}{these} multi-view features into the latent representations $\bm{\mathrm{A}}$, the aggregation loss is defined as $\mathcal{L}=\sum_{v=1}^{V}|m_{\bm{\theta}_v}(\bm{\mathrm{A}})-\bm{\mathit{f}}_{v}|^{2}$ \textcolor{black}{where} \textcolor{black}{$m_{\bm{\theta}_v}(\cdot)$} is a mapping function that \textcolor{black}{aims to project} the latent features $\bm{\mathrm{A}}$ to \textcolor{black}{the representation space of the $v$-$th$ view}.
\end{thm}

To measure the information loss between the latent representations $\bm{\mathrm{A}}$ and the original view features $\bm{\mathit{f}}_{v}$, we should transform them into a common representation space. Naturally, it is easy to consider projecting the view features $\bm{\mathit{f}}_{v}$ into the latent space \textcolor{black}{using the mapping function $m_{\bm{\theta}_v}(\cdot)$. However,} this makes the optimization process unstable. The latent space (i.e., the representation space of $\bm{\mathrm{A}}$) is unknown, which is the solution that we want to obtain. \textcolor{black}{Thus, we choose to measure the aggregation loss in the feature space of the original views. \textcolor{black}{It is well-known that each view consists of both the view-specific and the view-common information \cite{lyu2022beyond,yin2018multiview}}. Different from the explicit approaches \cite{lyu2022beyond,yin2018multiview}, our method leverages them in an implicit way. In the aggregation loss $\mathcal{L}$}, the squared $\ell_2$ norm is utilized to measure the distance between $\bm{\mathrm{A}}$ and $\big\{\bm{\mathit{f}}_{v}\big\}_{v=1}^{V}$. \textcolor{black}{The minimization of the aggregation loss $\mathcal{L}$ w.r.t. $\bm{\mathrm{A}}$ constrains that the original information in each view can be recovered from the latent representations, and thus the specific and the common information of views are implicitly aggregated into $\bm{\mathrm{A}}$ during the optimization process.} Fully exploiting the information of each view helps to build better data representations.

\vspace{0.3cm}
\noindent\textcolor{black}{$(b).$ \textit{Inverse Anchor Aggregator}}
\vspace{0.15cm}

\textcolor{black}{Our IAA consists of the two components which are the aggregation evaluator $\mathrm{Aggregation\_evaluator(\cdot|\bm{\mathrm{W}}, \bm{\mathrm{B}})}$ and the Adam optimizer}. The aggregation evaluator aims to measure the information loss during aggregating the anchors, while the Adam optimizer is used to optimize the learnable parameters of the aggregation evaluator or the representations of the latent anchors. \textcolor{black}{When using IAA to update the latent anchor representations, we first use the aggregation evaluator to measure the information loss between the latent anchors and the features of the original views as shown in Eq.\ref{eq:12},}
\begin{equation}
\begin{split}
\mathcal{L}=&\mathrm{Aggregation\_evaluator}(\bm{\mathrm{A}}, \bm{\mathrm{F}}^{S}|\bm{\mathrm{W}},\bm{\mathrm{B}})\\
=&\sum_{v=1}^{V}|\bm{w}_{v}\bm{\mathrm{A}}+\bm{b}_{v}-\bm{\mathrm{F}}_{v}^{S}|^{2}.
\end{split}
\label{eq:12}
\end{equation}
In this equation,  $\bm{\mathrm{A}}=\big\{\bm{\chi}_m\big\}_{m=1}^{N_S\times N_F}$ denotes the latent anchors. $\bm{w}_{v}\in \bm{\mathrm{W}}$ and $\bm{b}_{v}\in\bm{\mathrm{B}}$ represent the weights and the biases about the $v$-$th$ view observations, \textcolor{black}{which try to linearly transform the latent anchors $\bm{\mathrm{A}}$ to the original feature space of the $v$-$th$ view. Second, according to the aggregation loss $\mathcal{L}$, the representations of the latent anchors can be updated using the Adam optimizer, $\bm{\mathrm{A}}\gets~\bm{\mathrm{A}}-\nabla_{\bm{\mathrm{A}}}\mathcal{L}$. In particular, the labels of the latent anchors are represented as $\bm{\mathrm{Y}}=\big\{y_m\big\}_{m=1}^{N_S\times N_F}$, which are as same as the anchor labels before the aggregation, i.e., $\bm{\mathrm{Y}}=\bm{\mathrm{Y}}^{S}_{1}=...=\bm{\mathrm{Y}}^{S}_{V}$.} \par

\vspace{0.3cm}
\noindent\textcolor{black}{$(c).$ \textit{Workflow of Inverse Anchor Aggregation}}
\vspace{0.15cm}

\textcolor{black}{Our inverse anchor aggregation works in the two alternate steps, namely the aggregation evaluator update step and the latent anchor update step, which aim at progressively building a stronger aggregation evaluator while aggregating the Gaussian anchors into a unified latent space.}

\textcolor{black}{\textbf{Aggregation evaluator update step}}. This step is to help the aggregation evaluator to measure the aggregation loss  more accurately, \textcolor{black}{thereby facilitating the aggregation for the latent Gaussian anchors}. To realize this goal, we initialize the parameters of the aggregation evaluator using the Xavier initialization \cite{glorot2010understanding}, and further optimize these parameters to make it possible to project the latent anchors into the original view space as much as possible. We find that we can minimize the aggregation loss $\mathcal{L}$ w.r.t. \textcolor{black}{the learnable parameters $\bm{\mathrm{W}}$ and $\bm{\mathrm{B}}$ of the aggregation evaluator}, because correctly mapping $\bm{\mathrm{A}}$ to the original view space reduces their distance to the original view observations. Thus, the aggregation evaluator can be updated, $\bm{\mathrm{W}}\gets\bm{\mathrm{W}}-\nabla_{\bm{\mathrm{W}}}\mathcal{L}$ and $\bm{\mathrm{B}}\gets\bm{\mathrm{B}}-\nabla_{\bm{\mathrm{B}}}\mathcal{L}$, where $\nabla_{\bm{\mathrm{W}}}\mathcal{L}$ and $\nabla_{\bm{\mathrm{B}}}\mathcal{L}$ denote the gradient generated by the Adam optimizer according to the aggregation loss $\mathcal{L}$.

\textcolor{black}{\textbf{Latent anchor update step}}. \textcolor{black}{When the aggregation evaluator is updated, its parameters are frozen and it is utilized to guide the aggregation of the latent anchors}. Specifically, the aggregation evaluator is first used to assess the aggregation loss $\mathcal{L}$ \textcolor{black}{when aggregating the Gaussian anchors into the latent space, which is as same as the procedure shown in Eq.~\ref{eq:12}.} Then, we minimize the aggregation loss $\mathcal{L}$ w.r.t. the representations of the latent anchors~$\bm{\mathrm{A}}$. Therefore, the latent anchors can be updated as follows, $\bm{\mathrm{A}}\gets\bm{\mathrm{A}}-~\nabla_{\bm{\mathrm{A}}}\mathcal{L}$.  \par

\textcolor{black}{On one hand, the alternate work of these two steps can make it more effective for the aggregation evaluator to measure the aggregation loss.} On the other hand, the latent dense anchors become more representative to represent the original partial multi-view data as well, \textcolor{black}{thereby densely anchoring them into a unified latent space}. \textcolor{black}{The iteration number of this process is set as $iter=30$ in our work. Moreover, in each iteration, these two steps are both repeatedly conducted $N=10$ times to ensure the optimization stability.}\par

\subsubsection{Inverse Query Aggregation}
\textcolor{black}{We aggregate the queries into the latent space as well} by using the trained aggregation evaluator. \textcolor{black}{Note that} the parameters of the aggregation evaluator are frozen \textcolor{black}{in this stage. Specifically,} we first measure the information loss between the latent queries $\bm{\mathrm{H}}$ and the \textcolor{black}{multi-view query representations $\bm{\mathrm{F}}^{Q}$ using the aggregation evaluator, which is shown in Eq. \ref{eq:16}}
\begin{equation}
\begin{split}
\mathcal{L}'=&\mathrm{Aggregation\_evaluator}(\bm{\mathrm{H}}, \bm{\mathrm{F}}^{Q}|\bm{\mathrm{W}},\bm{\mathrm{B}})\\
=&\sum_{v=1}^{V}|\bm{w}_{v}\bm{\mathrm{H}}+\bm{b}_{v}-\bm{\mathrm{F}}_{v}^{Q}|^{2}.
\end{split}
\label{eq:16}
\end{equation}
Then, the latent representations of the queries $\bm{\mathrm{H}}$ are updated by using the Adam optimizer, $\bm{\mathrm{H}}\gets\bm{\mathrm{H}}-\nabla_{\bm{\mathrm{H}}}\mathcal{L}'$. By progressively minimizing the aggregation loss $\mathcal{L}'$ w.r.t. the latent queries $\bm{\mathrm{H}}$, the multi-view observations are unified into the latent representations $\bm{\mathrm{H}}=\big\{\bm{h}_n\big\}_{n=1}^{N_Q}$ in which the information of each view is fully considered.

\subsection{Anchor Distribution Self-Rectification}
\textcolor{black}{The Anchor Distribution Self-Rectification (ADSR) aims to further calibrate the latent anchors, therefore helping to build a more effective metric classifier. Specifically, we first calculate the central anchor representations for each of the categories $\bm{\mathrm{C}}$ using Eq. \ref{eq:23}},
\begin{equation}
\bm{\delta}_{c}=\frac{1}{|\bm{\mathrm{A}}_{c}|}\sum_{\bm{\chi}\in\bm{\mathrm{A}}_{c}} \bm{\chi}
\label{eq:23}
\end{equation}
in which $\bm{\mathrm{A}}_{c}$ represents the set of the anchors that belong to the class~$c$, \textcolor{black}{while the vector $\bm{\delta}_{c}\in \big\{\bm{\delta}_{c}\big\}_{c=1}^{|\bm{\mathrm{C}}|}$ denotes the central anchor representations of this class}.

\textcolor{black}{Afterwards, we measure the relation scores between the latent anchors $\bm{\mathrm{A}}$ and the categories $\bm{\mathrm{C}}$, represented as $\big\{\bm{\alpha}_{m}\big\}_{m=1}^{N_{S}\times N_{F}}$ where $\bm{\alpha}_{m}\in \mathbb{R}^{|\bm{\mathrm{C}}|\times 1}$ denotes the relation scores between \textcolor{black}{the latent anchor $\bm{\chi}_{m}$} and the categories $\bm{\mathrm{C}}$. The relations scores $\bm{\alpha}_{m}$ are calculated by the following two steps.} Firstly, we calculate the distance of $\bm{\chi}_{m}$ to each of the categories $\bm{\mathrm{C}}$ using Eq. \ref{eq:24},
\begin{equation}
\sigma_{m,c}=|\bm{\chi}_{m}-\bm{\delta}_c|^{2}
\label{eq:24}
\end{equation}
where $\sigma_{m,c}\in\bm{\sigma}_{m}$ denotes the distance between the anchor $\bm{\chi}_{m}$ and the category $c$, while $\bm{\sigma}_{m}=\big\{\sigma_{m,c}\big\}_{c=1}^{|\bm{\mathrm{C}}|}$ \textcolor{black}{represents the set of the distance between the anchor $\bm{\chi}_{m}$ and all of these $|\bm{\mathrm{C}}|$ classes}.
Secondly, the relation scores are drawn by using the following equation,
\begin{equation}
\bm{\alpha}_{m}=\mathrm{Softmax}(-\bm{\sigma}_{m}).
\label{eq:25}
\end{equation}
\textcolor{black}{According to these relation scores, the cross-entropy loss} $\mathcal{L}_{ce}$ is used to measure the correctness of the anchor distribution (the anchors should have lower distance to the classes that they belong to while being distant from the others),
\begin{equation}
\mathcal{L}_{ce}=-\frac{1}{N_{S}\times N_{F}}\sum_{m=1}^{N_{S}\times N_{F}}\sum_{c=1}^{|\bm{\mathrm{C}}|} y_{m,c}\log(\alpha_{m,c}),
\label{eq:26}
\end{equation}
\begin{equation}
s.t.\,\,\,\, \alpha_{m,c}\in\bm{\alpha}_{m}, \,\,\,\, y_{m,c}\in\bm{y}_{m}, \,\,\,\, \bm{y}_{m}=one\_hot(y_{m}).
\label{eq:27}
\end{equation}

\textcolor{black}{Also, the relation scores of the queries to the categories $\bm{\mathrm{C}}$} are measured as well following the equations Eq.~\ref{eq:24} and Eq.~\ref{eq:25}, denoted as $\big\{\bm{\beta}_{n}\big\}_{n=1}^{N_{Q}}$. We hope the latent anchors can correctly reflect the distribution of the queries. Thus, the Shannon entropy $\mathcal{L}_{se}$ is used to reduce the bias between the distribution of the anchors and the queries. Specifically, we first compute the global \textcolor{black}{query relation scores $\big\{\zeta_{c}\big\}_{c=1}^{|\bm{\mathrm{C}}|}$ about these $|\bm{\mathrm{C}}|$ categories, where} $\zeta_{c}$ is obtained by averaging the relation scores of all the queries to the category $c$,
\begin{equation}
\zeta_{c}=\frac{1}{N_Q}\sum_{n=1}^{N_{Q}} \beta_{n,c}, \quad s.t.\quad\beta_{n,c}\in \bm{\beta}_{n}.
\label{eq:30}
\end{equation}
We maximize the Shannon-entropy loss $\mathcal{L}_{se}$ w.r.t. the global query relation scores $\big\{\zeta_{c}\big\}_{c=1}^{|\bm{\mathrm{C}}|}$,
\begin{equation}
\mathcal{L}_{se}=\textcolor{black}{-}\sum_{c=1}^{|\bm{\mathrm{C}}|}\zeta_{c}\log(\zeta_{c}).
\label{eq:29}
\end{equation}
\textcolor{black}{As a result, the variables} $\big\{\zeta_{c}\big\}_{c=1}^{|\bm{\mathrm{C}}|}$ are constrained to obey the uniform distribution, thereby ensuring that the global relation between the queries and the anchors of different categories is equal as much as possible. Thus, the queries are guided to be not biased to the anchors of some specific categories during the rectification procedure. Equipping the cross-entropy term $\mathcal{L}_{ce}$ with the Shannon-entropy term $\mathcal{L}_{se}$ can realize better results as shown in Fig. \ref{fig:6}. \textcolor{black}{The total objective function of our ADSR can be described by Eq. \ref{eq:31},}
\begin{equation}
\big\{\bm{\delta}_{c}^{*}\big\}_{c=1}^{|\bm{\mathrm{C}}|}=\mathop{\arg\min}_{\big\{\bm{\delta}_{c}\big\}_{c=1}^{|\bm{\mathrm{C}}|}} (\mathcal{L}_{ce}\textcolor{black}{-}\mathcal{L}_{se})
\label{eq:31}
\end{equation}
where $\big\{\bm{\delta}_{c}^{*}\big\}_{c=1}^{|\bm{\mathrm{C}}|}$ denotes \textcolor{black}{the updated central anchor representations.}

Finally, the distribution offsets $\big\{\bm{\vartheta}_{c}\big\}_{c=1}^{|\bm{\mathrm{C}}|}$ for the latent anchors of \textcolor{black}{the categories $\bm{\mathrm{C}}$ are calculated by using the following equation},
\begin{equation}
\bm{\vartheta}_{c}=\bm{\delta}^{*}_{c}-\bm{\delta}_{c}.
\label{eq:33}
\end{equation}
\textcolor{black}{According to the distribution offsets}, \textcolor{black}{the representations of the latent anchors can be updated as below},
\begin{equation}
\bm{\chi}_{m}^{*} = \bm{\chi}_{m} + \bm{\vartheta}_{y_{m}}
\label{eq:34}
\end{equation}
\textcolor{black}{where $y_{m}$ and $\bm{\vartheta}_{y_{m}}$ denote the label and the distribution offset of the anchor $\bm{\chi}_{m}$, while $\bm{\chi}_{m}^{*}\in\bm{\mathrm{A}}^{*}$ represents the rectified anchor representations.} \textcolor{black}{Based on these rectified anchors $\bm{\mathrm{A}}^{*}$, an effective anchor-based classifier $\bm{\Phi}=\big\{\bm{\phi}_{c}\big\}_{c=1}^{|\bm{\mathrm{C}}|}$ can be constructed, where $\bm{\phi}_{c}$ denotes the classifier weights about the class $c$. $\bm{\phi}_{c}$ is imprinted by averaging the representations of the rectified anchors belonging to the class $c$,}
\begin{equation}
\bm{\phi}_{c} = \frac{1}{|\bm{\mathrm{A}}_{c}^{*}|}\sum_{\bm{\chi}^{*}\in \bm{\mathrm{A}}_{c}^{*}} \bm{\chi}^{*}
\label{eq:35}
\end{equation}
where $\bm{\mathrm{A}}_{c}^{*}$ denotes the set of the rectified anchors that belong to the class $c$. The label of the query sample $\bm{\mathrm{X}}^{Q}_{n}$ can be inferred by applying the anchor-based classifier to its latent representations $\bm{h}_{n}\in \bm{\mathrm{H}}$,
\begin{align}
y^{Q}_{n} & =\mathop{\arg\max}_{y'\in \bm{\mathrm{C}}} P (y'|\bm{\mathrm{X}}^{Q}_{n},\bm{\mathrm{S}})\\
&=\mathop{\arg\max}_{y'\in\bm{\mathrm{C}}}\frac{exp(-|\bm{\phi}_{y'}-\bm{h}_{n}|^{2})}{\sum_{c\in\bm{\mathrm{C}}}exp(-|\bm{\phi}_{c}-\bm{h}_{n}|^{2})}.
\label{eq:36}
\end{align}
\textcolor{black}{The pseudo codes of our unified Gaussian dense-anchoring method are provided in Algorithm 1}.

\begin{algorithm}[t]
\SetKwInOut{Input}{Input}\SetKwInOut{Output}{Output}
\SetAlgoLined
\caption{\textcolor{black}{The pseudo codes of our unified Gaussian dense-anchoring approach.}}
\Input{The supports $\bm{\mathrm{S}}=\big\{\bm{\mathrm{X}}^{S}_{n},y^{S}_{n}\big\}_{n=1}^{N_{S}}$, the queries $\bm{\mathrm{Q}}=\big\{\bm{\mathrm{X}}^{Q}_{n}\big\}_{n=1}^{N_{Q}}$, and the statistics of the base visual concepts $\left\{\big\{\ddot{\bm{\mu}}_{c,v}, \ddot{\bm{\varsigma}}_{c,v}\big\}_{c=1}^{|\bm{\mathrm{C}}_{b}|}\right\}_{v=1}^{V}$\;}
\Output{The label prediction of the queries $\big\{y^{Q}_{n}\big\}_{n=1}^{N_{Q}}$\;}
\textcolor{black}{Estimate the view distribution of the supports and the queries, $\bm{\Psi}^{S}$ and $\bm{\Psi}^{Q}$, through considering their available views and the statistics of the retrieved top-$k$ relevant base categories}\;
\textcolor{black}{Sample the dense Gaussian anchors $\bm{\mathrm{F}}^{S}$ from the support sample distribution}\;
\textcolor{black}{Interpolate the missing views of the queries, and update their feature representations $\bm{\mathrm{F}}^{Q}$}\;
$i\gets 1$, $j\gets 1$\;
\While{$i \leq iter$}{
  $t_1\gets 1$ \;
  \While{$t_1\leq N$}{
    \textcolor{black}{Calculate the aggregation loss $\mathcal{L}$}, $\mathcal{L}=\mathrm{Aggregation\_evaluator}(\bm{\mathrm{A}}, \bm{\mathrm{F}}^{S}|\bm{\mathrm{W}},\bm{\mathrm{B}})$\;
    \textcolor{black}{Update the learnable parameters of the aggregation evaluator}, $\bm{\mathrm{W}}\gets\bm{\mathrm{W}}-\nabla_{\bm{\mathrm{W}}}\mathcal{L}$ and $\bm{\mathrm{B}}\gets\bm{\mathrm{B}}-\nabla_{\bm{\mathrm{B}}}\mathcal{L}$\;
   $ t_1=t_1+1$\;
    }
$t_2\gets 1$ \;
\While{$t_2\leq N$}{
\textcolor{black}{Calculate the aggregation loss $\mathcal{L}$}, $\mathcal{L}=\mathrm{Aggregation\_evaluator}(\bm{\mathrm{A}}, \bm{\mathrm{F}}^{S}|\bm{\mathrm{W}},\bm{\mathrm{B}})$\;
\textcolor{black}{Update the latent anchors}, $\bm{\mathrm{A}}\gets\bm{\mathrm{A}}-\nabla_{\bm{\mathrm{A}}}\mathcal{L}$\;
$t_2=t_2+1$
}
$i=i+1$
}
\While{$j \leq iter$}{
    \textcolor{black}{Calculate the aggregation loss $\mathcal{L}'$},
    $\mathcal{L}'=\mathrm{Aggregation\_evaluator}(\bm{\mathrm{H}}, \bm{\mathrm{F}}^{Q}|\bm{\mathrm{W}},\bm{\mathrm{B}})$\;
    \textcolor{black}{Update the latent queries}, $\bm{\mathrm{H}}\gets\bm{\mathrm{H}}-\nabla_{\bm{\mathrm{H}}}\mathcal{L}'$\;
    $j=j+1$}
\textcolor{black}{Rectify the latent anchors $\bm{\mathrm{A}}$ by conducting the anchor distribution self-rectification}, $\bm{\mathrm{A}}^{*}\gets \mathrm{Rectify}(\bm{\mathrm{A}}, \bm{\mathrm{H}})$\;
\textcolor{black}{Build the anchor-base classifier $\bm{\Phi}=\big\{\bm{\phi}_{c}\big\}_{c=1}^{|\bm{\mathrm{C}}|}$ to categorize the queries}, $\big\{y^{Q}_{n}\big\}_{n=1}^{N_{Q}}=\mathrm{Classify}(\bm{\Phi}, \bm{\mathrm{H}})$\;
\end{algorithm}

\section{Experiments}
\textcolor{black}{The experiments are provided in this section to validate the effectiveness of our UGDA}. We first introduce the main datasets in Section 5.1. Then, we provide the implementation details in Section 5.2 and report the main experimental results in Section 5.3. Finally, in Section 5.4, we analyze our model in both qualitative and quantitative ways.

\subsection{\textcolor{black}{Main Datasets}}
Aiming to validate the effectiveness of our method, we have validated our UGDA on the six frequently used multi-view datasets that are Cub-googlenet-doc2vec \cite{zhang2019cpm}, Handwritten\footnote{\url{https://archive.ics.uci.edu/ml/datasets/Multiple+Features}}, Caltech102 \cite{li2015large,fei2004learning}, Scene15 \cite{fei2005bayesian}, Animal \cite{zhang2019cpm}, and ORL\footnote{\url{https://www.cl.cam.ac.uk/research/dtg/attarchive/facedatabase.html}}. Let us briefly introduce these datasets. \par

\textbf{Cub-googlenet-doc2vec}. \textcolor{black}{The Cub-googlenet-doc2vec dataset is collected from the images of the birds from the $10$ species. The deep visual features from GoogLeNet \cite{szegedy2015going} and the text features encoded by doc2vec \cite{le2014distributed} are utilized as the two views for each sample}. On the Cub-googlenet-doc2vec dataset, we randomly select the $6$ classes to build the base set and the $4$ classes to test the model.\par

\textbf{Handwritten}. The Handwritten dataset contains \textcolor{black}{the $2000$ images from the handwritten digits ``$0$'' $\sim$ ``$9$''. Also, the features of Fourier Coefficients, Profile Correlations, Karhunen-love Coefficients, Pixel Average in $2\times 3$ Windows, Zernike Moments, and Morphological are extracted from each image to form the six different views.} \textcolor{black}{We randomly choose} the $6$ categories from Handwritten to build the base set, while the rest $4$ categories are used for testing.\par

\textbf{Caltech102}. \textcolor{black}{The Caltech102 dataset totally contains the $102$ categories, including $101$ foreground object categories and a background category. For each sample, the features of Gabor, Wavelet Moments, CENTRIST, HOG, GIST, and LBP are extracted as the six views.} In the experiments, the $80$ categories of Caltech102 are randomly chosen to build the base set, while the rest $22$ categories are used for testing.\par

\textbf{Scene15}. The Scene15 dataset contains the $4485$ images collected from the $15$ different scenes. \textcolor{black}{The PHOG features and the GIST features are extracted from each image, which are treated as the two different views in this dataset. We randomly choose} the $8$ categories of Scene15 to build the base set, while the rest $7$ categories are used for testing. \par

\textbf{Animal}. The Animal dataset totally contains the $10158$ images collected from the animals of the $50$ categories. \textcolor{black}{The deep features extracted by DECAF \cite{krizhevsky2012imagenet} and VGG19 \cite{simonyan2014very} are treated as the two views in this dataset}. In our work, we randomly choose the $35$ categories to build the base set and the $15$ categories to test the model.\par

\textbf{ORL}. The ORL dataset consists of the $400$ facial samples \textcolor{black}{from} the $40$ subjects. \textcolor{black}{The features of Intensity, LBP, and Gabor are extracted from each sample in this dataset, which are treated as the three views.} In the experiments, the $25$ categories are randomly selected to build the base set, and the $15$ categories are used for testing. \par

\subsection{Implementation Details}
\textcolor{black}{We implement our codes using PyTorch with a TITAN RTX GPU card. We set the number of the sampled Gaussian anchors per support as $N_F=100$ in each view. Also, the hyper-parameter $k$ used in retrieving the neighboring base categories is set as $k=1$}.  Our inverse anchor aggregation works in an iterative manner. We set the iteration number as $iter=30$. \textcolor{black}{Moreover, in each iteration, the aggregation evaluator update step and the latent anchor update step are both repeatedly conducted $N=10$ times. The influence of these hyper-parameters on the model is studied in Section 5.4.3}. We use the Adam optimizer to optimize the learnable parameters of our UGDA, and set the initial learning rate as $1e^{-2}$. \textcolor{black}{According to the view-missing rate $\eta$, the partial multi-view data can be easily obtained from the conventional multi-view ones by randomly setting a part of the view indicators as zero to build the missing views.}  \textcolor{black}{Since our FPML is a new task, there are few appropriate methods that are specifically designed to tackle this challenge. Considering this issue}, we reimplement the six typical FSL methods (Proto \cite{snell2017prototypical}, Match \cite{vinyals2016matching}, TIM \cite{boudiaf2020information}, PTMAP \cite{hu2021leveraging}, RFS \cite{tian2020rethinking}, and BD-CSPN \cite{liu2020prototype}) and the four PML methods (CPM \cite{zhang2019cpm}, CPM-GAN \cite{zhang2020deep}, ICMSC \cite{wang2020icmsc}, and COMPLETER \cite{lin2021completer}). We try to use them to address the FPML task. For these methods, the missing views are padded with the mean features of the available views having the same label in the episode as much as possible.\par


\begin{table*}
    \small
    \centering
    \caption{\textcolor{black}{The experimental results on the Cub-googlenet-doc2vec dataset under the $3$-way $1$-shot $2$-view setting. In the table, ``Acc'' indicates the accuracy, ``$\eta$'' denotes the view-missing rate, and ``SE'' represents the standard error. Also, the first-place, the second-place, and the third-place results in each column are highlighted by the gray background \textcolor{gray!60}{$\blacksquare$}, \textcolor{gray!30}{$\blacksquare$}, and \textcolor{gray!15}{$\blacksquare$} respectively, which is the same for Table \ref{tab:2} $\sim$ Table \ref{tab:8}.}}
    \setlength{\tabcolsep}{3.8mm}{
    \begin{tabular}{c|cccccccc}
        \toprule
        \diagbox{Method}{Acc $\pm$ SE (\%)}{$\eta$} & $0$ & $0.1$ & $0.2$ & $0.3$ & $0.4$ & $0.5$ \\
        \midrule
        Proto \cite{snell2017prototypical} & $87.04\pm 0.53$ & $78.78\pm 0.58$ & $71.21\pm 0.60$ & $63.03\pm 0.53$ & $57.36\pm 0.50$ & $53.66\pm 0.45$ \\
        Match \cite{vinyals2016matching} & $89.47\pm 0.50$ & $81.76\pm 0.61$ & $74.75\pm 0.60$ & $65.66\pm 0.61$ & $57.33\pm 0.57$ &	$50.04\pm 0.50$ \\
        TIM	\cite{boudiaf2020information} & $94.38\pm 0.36$ & $87.03\pm 0.53$ & $76.53\pm 0.68$ & $66.16\pm 0.69$ & $57.52\pm 0.63$ & $52.71\pm 0.59$ \\
        PTMAP	 \cite{hu2021leveraging} & \cellcolor{gray!15}{$94.50\pm 0.37$} & \cellcolor{gray!15}{$90.45\pm 0.44$} & $85.98\pm 0.53$ & $76.96\pm 0.61$ &	$67.64\pm 0.59$ & $56.87\pm 0.51$ \\
        RFS \cite{tian2020rethinking} & $88.01\pm 0.51$ & $75.61\pm 0.83$ & $67.91\pm 0.83$ & $57.03\pm 0.71$ &	$51.67\pm 0.61$ & $48.34\pm 0.46$ \\
        BD-CSPN	\cite{liu2020prototype} & $76.91\pm 0.87$ & $60.53\pm 1.02$ & $50.85\pm 0.93$ & $42.18\pm 0.66$ & $39.36\pm 0.53$ & $36.58\pm 0.32$ \\
        \midrule
        \midrule
        CPM \cite{zhang2019cpm} & $83.49\pm 0.51$ & $72.57\pm 0.72$ & $67.07\pm 0.78$ & $57.40\pm 0.71$ & $51.39\pm 0.61$ & $47.62\pm 0.54$ \\
        CPM-GAN \cite{zhang2020deep} & $86.93\pm 0.50$ & $75.73\pm 0.72$ & $67.07\pm 0.74$ & 	$57.91\pm 0.68$ & $51.37\pm 0.59$ & $45.29\pm 0.48$ \\
        ICMSC	\cite{wang2020icmsc} & $93.13\pm 0.41$ & $86.97\pm 0.51$ & $82.94\pm 0.48$ & $75.51\pm 0.51$ &	$70.44\pm 0.48$ & $51.94\pm 0.46$ \\
        COMPLETER \cite{lin2021completer} & $87.65\pm 0.48$
  & $85.03\pm 0.56$ & $82.74\pm 0.59$ & $79.77\pm 0.63$ & $75.70\pm 0.65$ & $49.33\pm 0.51$ \\
        \midrule
        \midrule
        PTMAP$+$UGDA & \cellcolor{gray!30}{$95.42\pm 0.40$} & \cellcolor{gray!60}{$91.02\pm 0.62$} & \cellcolor{gray!60}{$88.34\pm 0.67$} & \cellcolor{gray!30}{$83.28\pm 0.80$} & \cellcolor{gray!60}{$81.76\pm 0.83$} & \cellcolor{gray!30}{$77.52\pm 0.92$} \\
        RFS$+$UGDA & $91.12\pm 0.46$ & $88.50\pm 0.55$ & \cellcolor{gray!15}{$86.62\pm 0.71$} & \cellcolor{gray!15}{$82.59\pm 0.68$} & \cellcolor{gray!30}{$80.37\pm 0.75$} & \cellcolor{gray!60}{$77.95\pm 0.77$} \\
       UGDA & \cellcolor{gray!60}{$95.59\pm 0.32$} & \cellcolor{gray!30}{$90.76\pm 0.53$} & \cellcolor{gray!30}{$87.72\pm 0.59$} & \cellcolor{gray!60}{$83.28\pm 0.73$} & \cellcolor{gray!15}{$80.16\pm 0.83$} & \cellcolor{gray!15}{$76.69\pm 0.87$} \\
        \bottomrule
    \end{tabular}}
    \label{tab:1}
\end{table*}

\begin{table*}
    \small
    \centering
    \caption{\textcolor{black}{The experimental results on the Handwritten dataset under the $3$-way $1$-shot $6$-view setting.}}
    \setlength{\tabcolsep}{3.8mm}{
    \begin{tabular}{c|cccccccc}
        \toprule
        \diagbox{Method}{Acc $\pm$ SE (\%)}{$\eta$} & $0$ & $0.1$ & $0.2$ & $0.3$ & $0.4$ & $0.5$ \\
        \midrule
        Proto \cite{snell2017prototypical} & $83.34\pm 0.54$ & $76.11\pm 0.59$ & $69.09\pm 0.55$ & $61.97\pm 0.58$ & $55.67\pm 0.49$ & $49.99\pm 0.45$ \\
        Match \cite{vinyals2016matching} & $83.39\pm 0.54$ & $77.03\pm 0.61$ & $70.09\pm 0.59$ & $62.53\pm 0.63$ & $55.82\pm 0.55$ &	$48.80\pm 0.46$ \\
        TIM	\cite{boudiaf2020information} & $87.56\pm 0.45$ & $79.40\pm 0.55$ & $70.65\pm 0.57$ & $61.94\pm 0.62$ & $53.76\pm 0.56$ & $46.66\pm 0.49$ \\
        PTMAP	 \cite{hu2021leveraging} & \cellcolor{gray!30}{$89.14\pm 0.44$} & \cellcolor{gray!15}{$83.15\pm 0.52$} & $74.91\pm 0.60$ & $66.70\pm 0.61$ &	$57.80\pm 0.63$ & $53.49\pm 0.64$ \\
        RFS \cite{tian2020rethinking} & $82.97\pm 0.54$ & $72.30\pm 0.75$ & $64.37\pm 0.73$ & $59.29\pm 0.68$ &	$54.99\pm 0.58$ & $51.90\pm 0.54$ \\
        BD-CSPN	\cite{liu2020prototype} & $87.39\pm 0.53$ & $70.32\pm 1.11$ & $56.55\pm 1.11$ & $48.06\pm0.96$ &	$41.24\pm 0.71$ & $36.63\pm 0.44$ \\
        \midrule
        \midrule
        CPM \cite{zhang2019cpm} & $83.48\pm 0.50$ & $69.88\pm 0.72$ & $61.20\pm 0.67$ & $55.04\pm 0.62$ & $50.82\pm 0.52$ & $46.27\pm 0.45$ \\
        CPM-GAN \cite{zhang2020deep} & $83.11\pm 0.53$ & $69.88\pm 0.72$ & $61.13\pm 0.67$ & 	$54.92\pm 0.62$ & $50.44\pm 0.51$ & $45.74\pm 0.42$ \\
        ICMSC	\cite{wang2020icmsc} & $84.51\pm 0.52$ & $72.66\pm 0.73$ & $63.03\pm 0.67$ & $53.81\pm 0.63$ &	$47.38\pm 0.52$ & $40.71\pm 0.52$ \\
        COMPLETER \cite{lin2021completer} & $84.38\pm 0.53$ & $78.11\pm 0.66$ & $71.51\pm 0.67$ & $66.52\pm 0.70$ & $62.76\pm 0.64$ & $58.58\pm 0.62$ \\
        \midrule
        \midrule
        PTMAP$+$UGDA & \cellcolor{gray!60}{$90.41\pm 0.49$} & \cellcolor{gray!30}{$85.23\pm 0.57$} & \cellcolor{gray!30}{$79.03\pm 0.68$} & \cellcolor{gray!15}{$72.51\pm 0.70$} & \cellcolor{gray!15}{$68.53\pm 0.70$} & \cellcolor{gray!30}{$65.54\pm 0.70$} \\
        RFS$+$UGDA & $85.55\pm 0.48$ & $81.44\pm 0.54$ & \cellcolor{gray!15}{$77.22\pm 0.52$} & \cellcolor{gray!30}{$73.52\pm 0.57$} & \cellcolor{gray!30}{$69.22\pm 0.58$} & \cellcolor{gray!15}{$64.42\pm 0.61$} \\
        UGDA & \cellcolor{gray!15}{$89.08\pm 0.42$} & \cellcolor{gray!60}{$85.44\pm 0.52$} & \cellcolor{gray!60}{$80.51\pm 0.57$} & \cellcolor{gray!60}{$76.13\pm 0.65$} & \cellcolor{gray!60}{$71.27\pm 0.66$} & \cellcolor{gray!60}{$66.77\pm 0.69$} \\
        \bottomrule
    \end{tabular}}
    \label{tab:2}
\end{table*}

\begin{table*}
    \small
    \centering
    \caption{\textcolor{black}{The experimental results on the Caltech102 dataset under the $5$-way $1$-shot $6$-view setting.}}
    \setlength{\tabcolsep}{3.8mm}{
    \begin{tabular}{c|cccccccc}
        \toprule
        \diagbox{Method}{Acc $\pm$ SE (\%)}{$\eta$} & $0$ & $0.1$ & $0.2$ & $0.3$ & $0.4$ & $0.5$ \\
        \midrule
        Proto \cite{snell2017prototypical} & $50.68\pm 0.52$ & $43.00\pm 0.57$ & $37.37\pm 0.51$ & $32.83\pm 0.42$ & $30.19\pm 0.35$ & $27.49\pm 0.29$ \\
        Match \cite{vinyals2016matching} & $53.64\pm 0.54$ & $48.13\pm 0.53$ & $43.49\pm 0.51$ & $38.85\pm 0.47$ & $34.40\pm 0.38$ &	$31.35\pm 0.34$ \\
        TIM	\cite{boudiaf2020information} & $56.48\pm 0.57$ & $45.75\pm 0.74$ & $39.27\pm 0.69$ & $33.31\pm 0.62$ & $28.33\pm 0.49$ & $25.57\pm 0.38$ \\
        PTMAP	 \cite{hu2021leveraging} & \cellcolor{gray!60}{$59.58\pm 0.52$} & \cellcolor{gray!30}{$53.49\pm 0.53$} & \cellcolor{gray!15}{$46.56\pm 0.50$} & $39.69\pm 0.46$ & $34.62\pm 0.37$ & $30.99\pm 0.33$ \\
        RFS \cite{tian2020rethinking} & $53.67\pm 0.52$ & $39.95\pm 0.72$ & $34.05\pm 0.65$ & $30.01\pm 0.49$ &	$27.24\pm 0.33$ & $26.08\pm 0.26$ \\
        BD-CSPN	\cite{liu2020prototype} & $47.23\pm 0.57$ & $34.24\pm 0.69$ & $28.04\pm 0.56$ & $23.49\pm 0.38$ &	$21.22\pm 0.22$ & $20.62\pm 0.14$ \\
        \midrule
        \midrule
        CPM \cite{zhang2019cpm} & $52.97\pm 0.52$ & $46.69\pm 0.52$ & $41.92\pm 0.49$ & $37.98\pm 0.46$ & $33.86\pm 0.37$ & $31.18\pm 0.34$ \\
        CPM-GAN \cite{zhang2020deep} & $52.67\pm 0.53$ & $46.63\pm 0.52$ & $42.11\pm 0.48$ & 	$38.03\pm 0.46$ & $34.36\pm 0.38$ & $31.46\pm 0.34$ \\
        ICMSC	\cite{wang2020icmsc} & $42.85\pm 0.53$ & $36.99\pm 0.49$ & $31.30\pm 0.40$ & $28.18\pm 0.35$ &	$24.72\pm 0.30$ & $22.35\pm 0.23$ \\
        COMPLETER \cite{lin2021completer} & $47.39\pm 0.55$ & $42.64\pm 0.54$ & $39.31\pm 0.53$ & $34.54\pm 0.45$ & $31.11\pm 0.38$ & $28.74\pm 0.33$ \\
        \midrule
        \midrule
        PTMAP$+$UGDA &$56.58\pm 0.61$ & \cellcolor{gray!15}{$53.45\pm 0.62$} & \cellcolor{gray!30}{$50.21\pm 0.58$} & \cellcolor{gray!30}{$46.86\pm 0.58$} & \cellcolor{gray!30}{$43.35\pm 0.55$} & \cellcolor{gray!30}{$40.76\pm 0.55$} \\
        RFS$+$UGDA & \cellcolor{gray!15}{$57.24\pm 0.54$} & $51.37\pm 0.58$ & $46.30\pm 0.60$ & \cellcolor{gray!15}{$42.29\pm 0.54$} & \cellcolor{gray!15}{$39.88\pm 0.50$} & \cellcolor{gray!15}{$37.22\pm 0.43$} \\
        UGDA & \cellcolor{gray!30}{$59.13\pm 0.59$} & \cellcolor{gray!60}{$54.15\pm 0.60$} & \cellcolor{gray!60}{$51.35\pm 0.57$} & \cellcolor{gray!60}{$47.81\pm 0.56$} & \cellcolor{gray!60}{$44.29\pm 0.53$} & \cellcolor{gray!60}{$42.05\pm 0.52$} \\
        \bottomrule
    \end{tabular}}
    \label{tab:3}
\end{table*}

\begin{table*}
    \small
    \centering
    \caption{\textcolor{black}{The experimental results on the Scene15 dataset under the $3$-way $1$-shot $2$-view setting.}}
    \setlength{\tabcolsep}{3.8mm}{
    \begin{tabular}{c|cccccccc}
        \toprule
        \diagbox{Method}{Acc $\pm$ SE (\%)}{$\eta$} & $0$ & $0.1$ & $0.2$ & $0.3$ & $0.4$ & $0.5$ \\
        \midrule
        Proto \cite{snell2017prototypical} & $67.71\pm 0.69$ & $58.74\pm 0.66$ & $52.09\pm 0.61$ & $47.34\pm 0.54$ & $44.72\pm 0.50$ & $44.27\pm 0.46$ \\
        Match \cite{vinyals2016matching} & $68.21\pm 0.71$ & $58.05\pm 0.68$ & $51.49\pm 0.61$ & $46.32\pm 0.54$ & $44.08\pm 0.51$ &	$43.04\pm 0.45$ \\
        TIM	\cite{boudiaf2020information} & \cellcolor{gray!60}{$74.21\pm 0.87$} & $63.19\pm 0.85$ & $53.01\pm 0.72$ & $46.78\pm 0.66$ & $45.07\pm 0.67$ & $45.34\pm 0.62$ \\
        PTMAP	 \cite{hu2021leveraging} & $70.38\pm 0.68$ & $66.72\pm 0.68$ & $63.49\pm 0.66$ & $58.77\pm 0.63$ &	$53.79\pm 0.58$ & $50.68\pm 0.52$ \\
        RFS \cite{tian2020rethinking} & $67.56\pm 0.66$ & $64.51\pm 0.64$ & $60.66\pm 0.65$ & $57.59\pm 0.58$ &	$52.66\pm 0.54$ & $49.76\pm 0.50$ \\
        BD-CSPN \cite{liu2020prototype} &$50.99\pm 0.83$ & $46.73\pm 0.70$ & $42.84\pm 0.62$ & $39.06\pm 0.46$ &	$36.91\pm 0.37$ & $35.58\pm 0.26$ \\
        \midrule
        \midrule
        CPM \cite{zhang2019cpm} & $55.05\pm 0.59$ & $50.64\pm 0.59$ & $48.29\pm 0.54$ & $44.11\pm 0.49$ & $41.80\pm 0.45$ & $39.66\pm 0.43$ \\
        CPM-GAN \cite{zhang2020deep} & $53.03\pm 0.59$ & $46.62\pm 0.54$ & $46.97\pm 0.52$ & 	$45.51\pm 0.51$ & $42.79\pm 0.44$ & $39.84\pm 0.44$ \\
        ICMSC	\cite{wang2020icmsc} & $67.28\pm 0.71$ & $58.31\pm 0.68$ & $52.39\pm 0.62$ & $47.82\pm 0.62$ &	$44.09\pm 0.51$ & $42.14\pm 0.45$ \\
        COMPLETER \cite{lin2021completer} & $69.34\pm 0.69$ & $65.55\pm 0.67$ & $61.63\pm 0.65$ & $58.37\pm 0.63$ & $53.93\pm 0.60$ & $43.20\pm 0.48$ \\
        \midrule
        \midrule
        PTMAP$+$UGDA & \cellcolor{gray!15}{$71.74\pm 0.76$} & \cellcolor{gray!30}{$69.63\pm 0.76$} & \cellcolor{gray!30}{$67.09\pm 0.79$} & \cellcolor{gray!30}{$65.46\pm 0.75$} &	\cellcolor{gray!30}{$62.90\pm 0.77$} & \cellcolor{gray!60}{$62.46\pm 0.80$} \\
        RFS$+$UGDA & $70.60\pm 0.68$ & \cellcolor{gray!15}{$68.65\pm 0.67$} & \cellcolor{gray!15}{$65.66\pm 0.68$} & \cellcolor{gray!15}{$63.92\pm 0.65$} & \cellcolor{gray!15}{$61.35\pm 0.68$} & \cellcolor{gray!15}{$60.25\pm 0.71$} \\
        UGDA & \cellcolor{gray!30}{$72.37\pm 0.65$} & \cellcolor{gray!60}{$70.15\pm 0.63$} & \cellcolor{gray!60}{$67.11\pm 0.66$} & \cellcolor{gray!60}{$65.83\pm 0.66$} & \cellcolor{gray!60}{$63.26\pm 0.67$} & \cellcolor{gray!30}{$62.17\pm 0.69$} \\
        \bottomrule
    \end{tabular}}
    \label{tab:4}
\end{table*}

\begin{table*}
\small
    \centering
    \caption{\textcolor{black}{The experimental results on the Animal dataset under the $5$-way $1$-shot $2$-view setting.}}
    \setlength{\tabcolsep}{3.8mm}{
    \begin{tabular}{c|cccccccc}
        \toprule
        \diagbox{Method}{Acc $\pm$ SE (\%)}{$\eta$} & $0$ & $0.1$ & $0.2$ & $0.3$ & $0.4$ & $0.5$ \\
        \midrule
        Proto \cite{snell2017prototypical} & $83.91\pm 0.46$ & $64.71\pm 0.77$ & $53.98\pm 0.58$ & $49.08\pm 0.49$ & $47.00\pm 0.44$ & $44.94\pm 0.39$ \\
        Match \cite{vinyals2016matching} & $83.63\pm 0.46$ & $77.75\pm 0.48$ & $70.31\pm 0.49$ & $62.67\pm 0.50$ & $53.16\pm 0.48$ &	$43.40\pm 0.39$ \\
        TIM \cite{boudiaf2020information} & $86.61\pm 0.61$ & $80.04\pm 0.77$ & $70.82\pm 0.85$ & $61.27\pm 0.87$ & $50.90\pm 0.77$ & $40.85\pm 0.63$ \\
        PTMAP	 \cite{hu2021leveraging} & \cellcolor{gray!15}{$87.15\pm 0.42$} & \cellcolor{gray!15}{$81.70\pm 0.49$} & $74.82\pm 0.47$ & $67.42\pm 0.50$ &	$57.80\pm 0.47$ & $47.02\pm 0.41$ \\
        RFS \cite{tian2020rethinking} & $84.00\pm 0.45$ & $69.44\pm 0.69$ & $59.97\pm 0.57$ & $55.08\pm 0.50$ & $51.59\pm 0.46$ & $48.21\pm 0.40$ \\
        BD-CSPN	 & $82.11\pm 0.52$ & $50.17\pm 1.12$ & $39.77\pm 0.78$ & $39.47\pm 0.60$ &	$42.02\pm 0.60$ & $39.35\pm 0.50$ \\
        \midrule
        \midrule
        CPM \cite{zhang2019cpm} & $58.08\pm 0.52$ & $52.48\pm 0.51$ & $49.29\pm 0.49$ & $46.23\pm 0.48$ & $48.17\pm 0.44$ & $41.40\pm 0.35$ \\
        CPM-GAN \cite{zhang2020deep} & $59.43\pm 0.54$ & $53.93\pm 0.51$ & $52.07\pm 0.49$ & 	$47.27\pm 0.48$ & $47.21\pm 0.45$ & $39.50\pm 0.32$ \\
        ICMSC	\cite{wang2020icmsc} & $77.43\pm 0.55$ & $71.70\pm 0.58$ & $67.01\pm 0.57$ & $60.83\pm 0.55$ &	$55.62\pm 0.53$ & $23.73\pm 0.24$ \\
        COMPLETER \cite{lin2021completer} & $71.30\pm 0.56$ & $68.23\pm 0.56$ & $65.97\pm 0.56$ & $63.77\pm 0.58$ & $58.74\pm 0.58$ & $41.15\pm 0.38$ \\
        \midrule
        \midrule
        PTMAP$+$UGDA & \cellcolor{gray!60}{$92.23\pm 0.37$} & \cellcolor{gray!60}$86.49\pm 0.46$ & \cellcolor{gray!60}{$80.48\pm 0.54$}  & \cellcolor{gray!60}{$74.63\pm 0.60$} & \cellcolor{gray!60}{$68.26\pm 0.63$} & \cellcolor{gray!60}{$63.68\pm 0.64$} \\
        RFS$+$UGDA & $85.87\pm 0.43$ & $81.17\pm 0.48$ & \cellcolor{gray!15}{$76.42\pm 0.51$} & \cellcolor{gray!15}{$71.38\pm 0.53$} & \cellcolor{gray!30}{$66.72\pm 0.57$} & \cellcolor{gray!30}{$63.32\pm 0.53$} \\
        UGDA & \cellcolor{gray!30}{$89.86\pm 0.41$} & \cellcolor{gray!30}{$84.34\pm 0.47$} & \cellcolor{gray!30}{$77.96\pm 0.55$} & \cellcolor{gray!30}{$72.59\pm 0.57$} & \cellcolor{gray!15}{$66.71\pm 0.62$} & \cellcolor{gray!15}{$62.25\pm 0.64$} \\
        \bottomrule
    \end{tabular}}
    \label{tab:5}
\end{table*}

\begin{table*}
    \centering
    \small
    \caption{\textcolor{black}{The experimental results on the ORL dataset under the $5$-way $1$-shot $3$-view setting.}}
    \setlength{\tabcolsep}{3.8mm}{
    \begin{tabular}{c|cccccccc}
        \toprule
        \diagbox{Method}{Acc $\pm$ SE (\%)}{$\eta$} & $0$ & $0.1$ & $0.2$ & $0.3$ & $0.4$ & $0.5$ \\
        \midrule
        Proto \cite{snell2017prototypical} & $85.12\pm 0.46$ & $78.34\pm 0.49$ & $68.71\pm 0.54$ & $59.36\pm 0.63$ & $50.96\pm 0.45$ &$44.68\pm 0.43$ \\
        Match \cite{vinyals2016matching} & $85.56\pm 0.44$ & $77.21\pm 0.59$ & $69.33\pm 0.61$ & $60.21\pm 0.62$ & $51.30\pm 0.57$ &	$43.30\pm 0.50$ \\
        TIM	\cite{boudiaf2020information} & $84.66\pm 0.67$ & $77.82\pm 0.63$ & $66.90\pm 0.71$ & $56.73\pm  0.71$ & $47.21\pm 0.64$ & $45.45\pm 0.57$ \\
        PTMAP	 \cite{hu2021leveraging} & \cellcolor{gray!15}{$91.52\pm 0.34$} & \cellcolor{gray!15}{$87.99\pm 0.39$} & \cellcolor{gray!15}{$85.99\pm 0.39$} & $74.59\pm 0.45$ &	$64.99\pm 0.48$ & $55.68\pm 0.50$ \\
        RFS \cite{tian2020rethinking} & $86.51\pm 0.44$ & $80.04\pm  0.57$ & $71.59\pm 0.64$ & $62.13\pm 0.64$ &	$53.33\pm 0.58$ & $45.53\pm 0.46$ \\
        BD-CSPN & $74.18\pm 0.74$ & $51.61\pm 1.05$ & $37.37\pm 0.79$ & $28.73\pm 0.58$ &	$25.35\pm 0.45$ & $23.34\pm 0.33$ \\
        \midrule
        \midrule
        CPM \cite{zhang2019cpm} & $84.00\pm 0.45$ & $74.52\pm 0.58$ & $63.62\pm 0.63$ & $56.47\pm 0.64$ & $50.69\pm 0.53$ & $46.43\pm 0.49$ \\
        CPM-GAN \cite{zhang2020deep} & $83.77\pm 0.67$ & $74.34\pm 0.57$ & $63.61\pm 0.62$ & 	$55.58\pm 0.63$ & $50.19\pm 0.52$ & $46.57\pm 0.46$ \\
        ICMSC	\cite{wang2020icmsc} & $88.21\pm 0.43$ & $57.52\pm 1.51$ & $36.40\pm 1.26$ & $29.07\pm 0.95$ &	$22.69\pm 0.53$ & $20.02\pm 0.41$ \\
        COMPLETER \cite{lin2021completer} & $80.04\pm 0.55$ & $76.83\pm 0.57$ & $72.62\pm 0.59$ & $67.62\pm 0.58$ & $60.89\pm 0.57$ & $54.21\pm 0.53$ \\
        \midrule
        \midrule
        PTMAP$+$UGDA & \cellcolor{gray!60}{$96.95\pm 0.26$} & \cellcolor{gray!30}{$92.27\pm 0.40$} & \cellcolor{gray!30}{$86.02\pm 0.48$} &  \cellcolor{gray!60}{$79.17\pm 0.55$}  & \cellcolor{gray!30}{$71.15\pm 0.60$} & \cellcolor{gray!30}{$62.70\pm 0.63$} \\
        RFS$+$UGDA & $90.95\pm 0.35$ & $87.88\pm 0.42$ & $82.75\pm 0.47$ & \cellcolor{gray!15}{$76.96\pm 0.49$} & \cellcolor{gray!15}{$69.50\pm 0.53$} & \cellcolor{gray!15}{$62.58\pm 0.55$} \\
        UGDA & \cellcolor{gray!30}{$95.79\pm 0.34$} & \cellcolor{gray!60}{$92.64\pm 0.44$} & \cellcolor{gray!60}{$86.59\pm 0.55$} & \cellcolor{gray!60}{$81.41\pm 0.58$} & \cellcolor{gray!60}{$72.44\pm 0.61$} & \cellcolor{gray!60}{$64.78\pm 0.66$} \\
        \bottomrule
    \end{tabular}}
    \label{tab:6}
\end{table*}

\subsection{Main Experimental Results}
Our method is evaluated on the six multi-view datasets, of which the results are shown in Table~\ref{tab:1} $\sim$ Table~\ref{tab:6}. According to these results, we have the following observations. \textcolor{black}{Compared with the FSL methods, our UGDA obviously has better robustness to the view-missing issue. Take the experiments on the Cub-googlenet-doc2vec dataset as an example. When the view-missing rate $\eta$ increases to $0.5$, the accuracy of all the compared FSL methods decreases to around $50 \%$ and the accuracy drops all reach around $40 \%$. In contrast}, UGDA's accuracy can still be maintained at $76.69 \%$ and the accuracy drop is only $18.90 \%$. We find even at the view-missing rate $\eta=0$, UGDA can still achieve competitive performance, \textcolor{black}{e.g., its accuracy is obviously higher than that of Proto, Match, RFS, and BD-CSPN. Similar results can also be found in other datasets. Our UGDA is better than the compared PML methods as well. On one hand, \textcolor{black}{UGDA has better few-shot classification capability. Without the influence of the view-missing issue ($\eta=0$), the accuracy of our UGDA is consistently higher than that of CPM, CPM-GAN, ICMSC, and COMPLETER on all these six multi-view datasets. On the other hand}}, \textcolor{black}{in the scenario that the views are heavily missing, our method can still achieve higher accuracy due to its better robustness to \textcolor{black}{the view-missing issue}}. For example, the accuracy of UGDA is $22.51 \%$, $22.33 \%$, $20.03\%$, and $18.97 \%$ higher than that of CPM, CPM-GAN, ICMSC, and COMPLETER at $\eta=0.5$ on the Scene15 dataset. The above experimental results demonstrate the superiority of our approach.

\begin{figure*}[t!]
\centering
 \includegraphics[height=10cm]{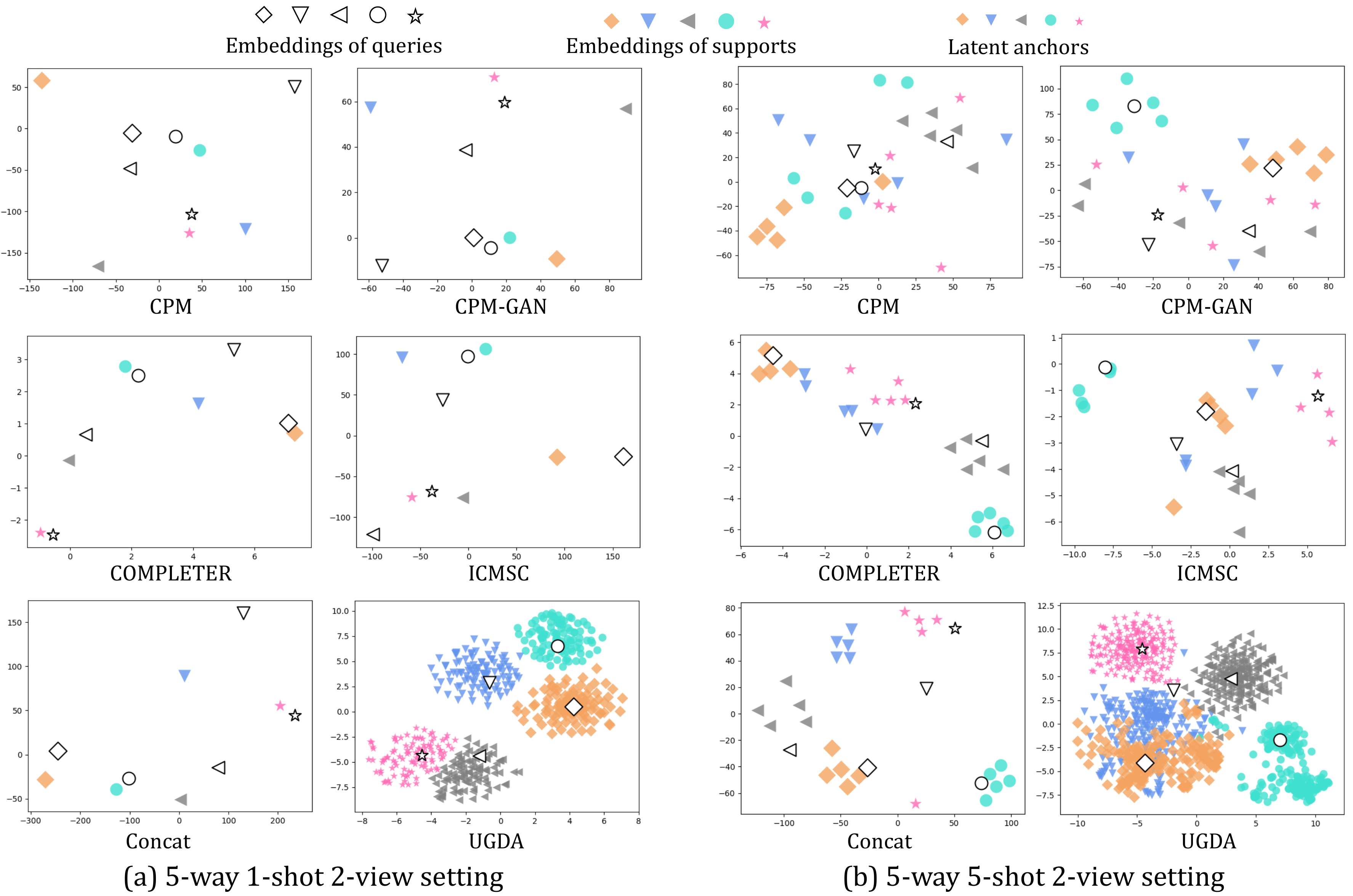}
\caption{\textcolor{black}{The visualized comparison between the different methods. The PML methods CPM, CPM-GAN, COMPLETER, and ICMSC infer the categories of the queries according to the support features. Thus, for these methods, we visualize the features of the supports and the queries learned in the subspace. In particular, the FSL methods Proto, Match, TIM, RFS, and BD-CSPN are not originally designed for processing multi-view data. Therefore, for these methods, we all concatenate the observations of different views to form the unified data representations, which is shown in the experiment ``Concat''. For our UGDA, we visualize the embeddings of the queries and the Gaussian anchors aggregated in the latent representation space. The above experiments are conducted on the Animal dataset under the view-missing rate $\eta=0.3$.}}
\label{fig:3}
\end{figure*}

\textcolor{black}{We also incorporate our UGDA with the FSL models PTMAP and RFS to demonstrate its compatibility}. \textcolor{black}{The experimental results indicate that UGDA can improve their robustness against the view-missing issue.} \textcolor{black}{We take the results on the Cub-googlenet-doc2vec dataset as an example. \textcolor{black}{At the view-missing rate $\eta=0.3$, the use of UGDA improves the accuracy of PTMAP from $76.96 \%$ to $83.28 \%$,} while the accuracy of RFS from $57.03 \%$ to $82.59 \%$. Correspondingly, at the view-missing rate $\eta=0.5$, UGDA can bring $20.65 \%$ and $29.61 \%$ accuracy improvement to PTMAP and RFS \textcolor{black}{respectively}.} We find that \textcolor{black}{even at the view-missing rate $\eta=0$}, our UGDA can still make these two models more accurate as fully exploiting the information of each view makes data more representative. \textcolor{black}{We qualitatively analyze the differences between the compared methods and our UGDA in Fig.~\ref{fig:3} using tSNE \cite{van2008visualizing}. On one hand, the concatenation is an intuitive and straightforward way to connect the observations of different views. But it cannot help to achieve the effective categorization of the queries, as view missing severely damages the consistency of the data while data scarcity \textcolor{black}{also} makes it harder to model the query distribution.} On the other hand, CPM, CPM-GAN, COMPLETER, and ICMSC cannot learn the effective unified representations from the limited partial multi-view data as well. The data-scarcity issue limits the effectiveness of these methods. In contrast, our dense Gaussian anchors can reflect the distribution of the queries more clearly, \textcolor{black}{which is beneficial for achieving robust and accurate classification results.}

\begin{figure*}[t!]
\centering
\includegraphics[height=9cm]{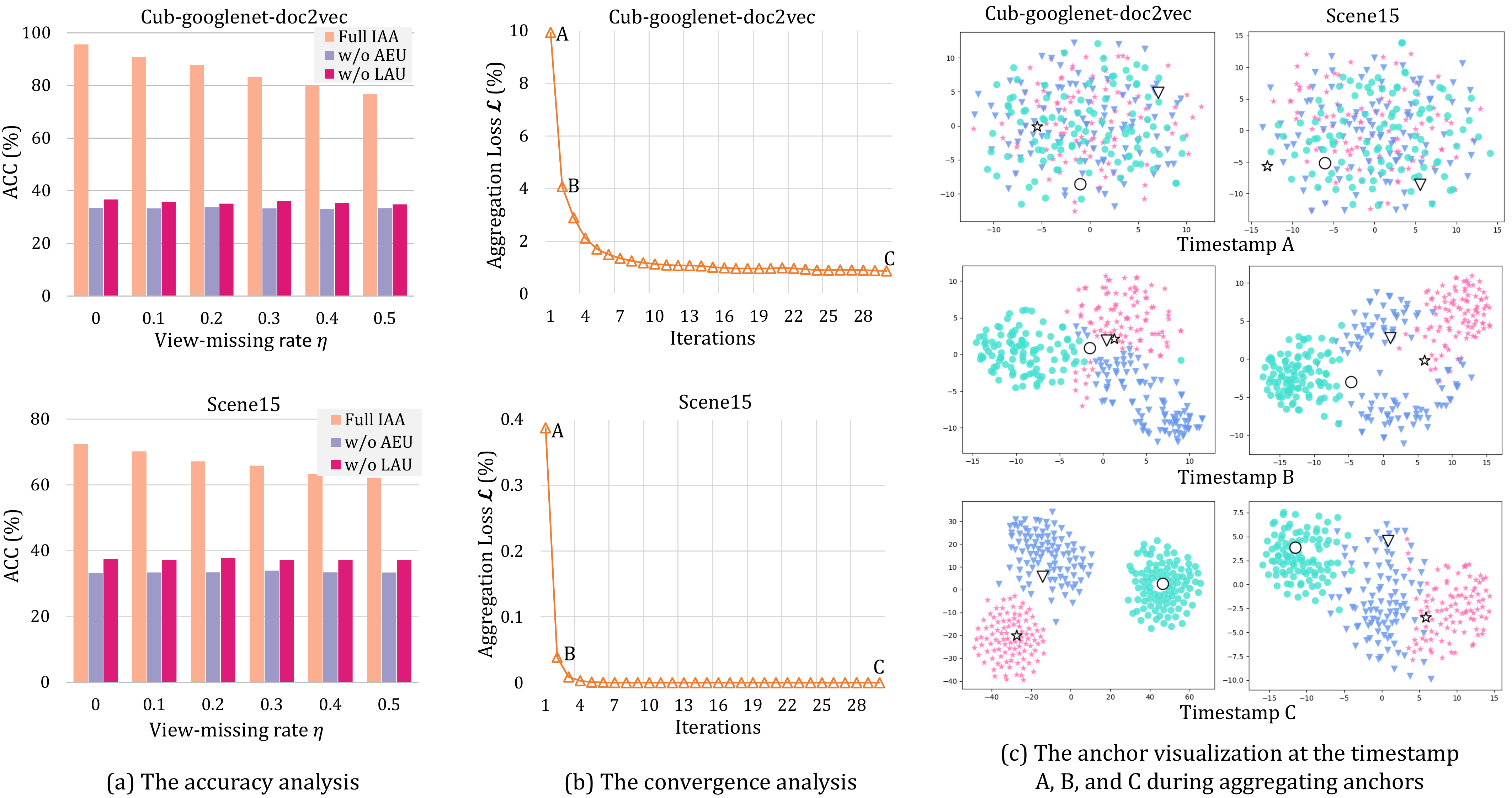}
\caption{\textcolor{black}{The quantitative and qualitative analysis for the inverse anchor aggregator. The above experiments are conducted on the Cub-googlenet-doc2vec and the Scene15 dataset under the 3-way 1-shot 2-view setting. Note that the timestamps A, B, and C are marked in the figure (b). ``AEU'' indicates the Aggregation Evaluator Update step and ``LAU'' denotes the Latent Anchor Update step. In the figure (c), the latent anchors are represented as the small coloured marks, while the embeddings of the queries are represented by the big white marks with black edge lines. The experiments (b) and (c) are conducted at $\eta=0.5$.}}
\label{fig:4}
\end{figure*}

\begin{figure*}[t!]
\centering
 \includegraphics[height=4.8cm]{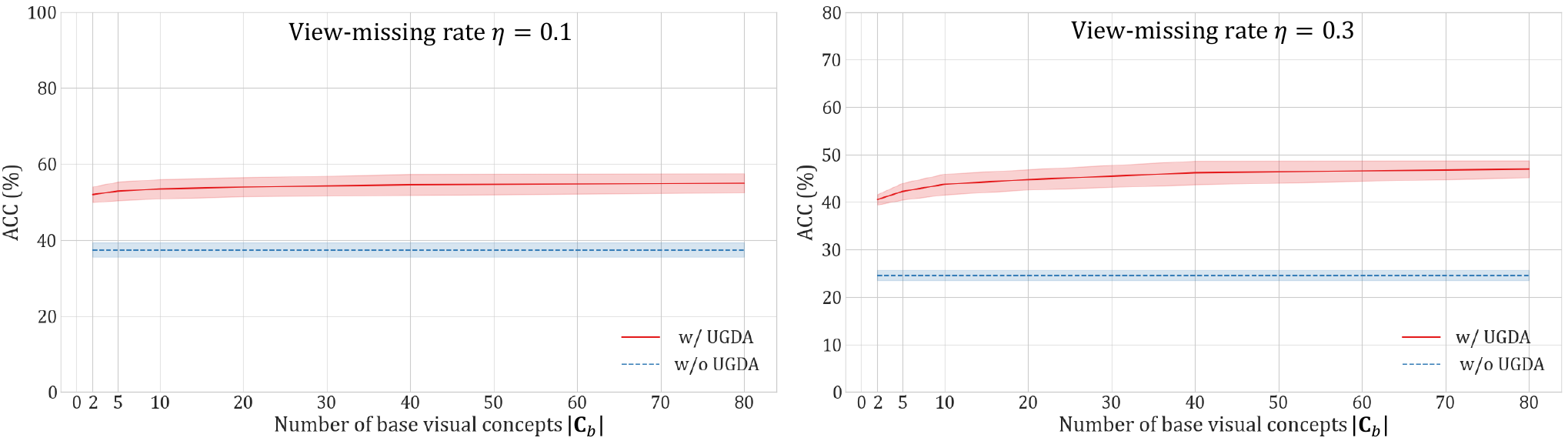}
\caption{\textcolor{black}{The robustness analysis for the number of the base visual concepts. The experiments are conducted on the Caltech102 dataset under the view-missing rate $\eta=0.1$ and $\eta=0.3$. Notice that the model is tested $5$ times in each experiment. In each time, the accuracy is evaluated based on the results of $400$ FPML tasks, and the base categories are randomly selected to build the base set. In  the experiments ``w/o UGDA'',  the multi-view observations are directly concatenated instead of being processed by our UGDA. In the figure, the light blue and the light red regions indicate the standard deviation}.}
\label{fig:5}
\end{figure*}

\begin{table*}
    \centering
    \caption{\textcolor{black}{The robustness analysis in the scenario that the base and the test categories are dissimilar. The experiments are conducted on the tieredImagenet dataset under the $5$-way $1$-shot $3$-view setting. In particular, we strictly follow the dataset split used in \cite{ren2018meta,he2020memory,tian2020rethinking} to ensure that the base and the test set are distinct enough. In the table, ``SE'' represents the standard error and ``$\eta$'' denotes the view-missing rate.}}
    \setlength{\tabcolsep}{4mm}{
    \begin{tabular}{c|cccccccc}
        \toprule
        \diagbox{Method}{Acc $\pm$ SE (\%)}{$\eta$} & $0.1$ & $0.2$ & $0.3$ & $0.4$ & $0.5$ & $0.6$ \\
        \midrule
        PTMAP \cite{hu2021leveraging} & \cellcolor{gray!30}{$73.90\pm 0.27$}  & \cellcolor{gray!15}{$70.69\pm 0.31$} & $66.28\pm 0.22$ & $60.34\pm 0.23$  & $53.19\pm 0.18$  & $44.12\pm 0.22$\\
        RFS  \cite{tian2020rethinking} & $62.04\pm 0.35$  & $53.79\pm 0.49$ & $46.67\pm 0.33$ & $40.21\pm 0.26$  & $37.31\pm 0.13$  & $34.81\pm 0.15$ \\
        ICMSC \cite{wang2020icmsc} & $57.23\pm 0.30$  & $46.74\pm 0.40$ & $38.91\pm 0.21$ & $32.06\pm 0.25$  & $26.34\pm 0.26$  & $23.29\pm 0.17$ \\
        COMPLETER \cite{lin2021completer} & $66.23\pm 0.38$  & $61.90\pm 0.47$ & $57.20\pm 0.31$ & $50.96\pm 0.21$  & $46.22\pm 0.27$  & $40.13\pm 0.25$ \\
        \midrule
        \midrule
        PTMAP+UGDA & \cellcolor{gray!15}{$73.76\pm 0.39$}  & \cellcolor{gray!30}{$71.68\pm 0.36$} & \cellcolor{gray!30}{$69.51\pm 0.32$} & \cellcolor{gray!30}{$66.92\pm 0.51$}  & \cellcolor{gray!30}{$64.64\pm 0.43$}  & \cellcolor{gray!60}{$61.86\pm 0.46$} \\
        RFS+UGDA & $70.67\pm 0.36$  & $68.82\pm 0.49$ & \cellcolor{gray!15}{$66.85\pm 0.41$} & \cellcolor{gray!15}{$64.47\pm 0.55$} & \cellcolor{gray!15}{$62.29\pm 0.42$} & \cellcolor{gray!15}{$59.48\pm 0.45$} \\
        UGDA & \cellcolor{gray!60}{$73.98\pm 0.36$}  & \cellcolor{gray!60}{$72.13\pm 0.41$} & \cellcolor{gray!60}{$69.65\pm 0.32$} & \cellcolor{gray!60}{$67.01\pm 0.47$}  & \cellcolor{gray!60}{$64.71\pm 0.41$}  & \cellcolor{gray!30}{$61.79\pm 0.47$} \\
        \bottomrule
    \end{tabular}}
    \label{tab:7}
\end{table*}

\subsection{\textcolor{black}{Quantitative and Qualitative Method Analysis}}
\subsubsection{\textcolor{black}{Analysis for Our Dense Gaussian Anchor Inverse-aggregation}}
\vspace{0.3cm}
\noindent\textcolor{black}{$(a).$ \textit{Ablation Study for Inverse Anchor Aggregator}}
\vspace{0.15cm}

\textcolor{black}{In this part, we conduct the ablation study for our Inverse Anchor Aggregator (IAA) used in the Dense Gaussian Anchor Inverse-aggregation (DGAI) phase. The experimental results are summarized in Fig.~\ref{fig:4}.} \textcolor{black}{In order to clearly reflect the influence of IAA,} the ablation study is conducted in both quantitative and qualitative ways.

\textbf{Quantitative analysis}. \textcolor{black}{The results shown in Fig.~\ref{fig:4} (a) indicate that the use of the full IAA can yield the best performance}. When the Aggregation Evaluator Update step or the Latent Anchor Update step is not employed (i.e., ``w/o AEU'' or ``w/o LAU''), the model performance drops obviously, \textcolor{black}{thus showing the importance of these two steps}. In addition, we also analyze the convergence of our inverse anchor aggregation, which is shown in Fig.~\ref{fig:4}~(b).  \textcolor{black}{The results indicate that} the aggregation loss $\mathcal{L}$ can be minimized stably with the increase of the iteration number. \textcolor{black}{As our inverse anchor aggregation} has good convergence, we empirically set the iteration number as $iter=30$ in this work. \par

\textbf{Qualitative analysis}. \textcolor{black}{For better reflecting the influence of IAA in aggregating anchors, we visualize the latent anchors at the timestamps A, B, and C using tSNE \cite{van2008visualizing}. Notice that the timestamps A, B, and C are marked at the curve of Fig. \ref{fig:4} (b).} From Fig. \ref{fig:4} (c), we can see that at the beginning, the latent anchors cannot clearly reflect the distribution of \textcolor{black}{the} queries. However, \textcolor{black}{when aggregating the available view information into the latent space, they gradually become more representative, e.g.}, the anchors \textcolor{black}{belonging} to the same category are clustered together while the anchors of the different categories are distant. Also, the latent anchors can reflect the categories of the queries more accurately.

\vspace{0.3cm}
\noindent\textcolor{black}{$(b).$ \textit{Robustness Analysis for Base Visual Concept Number}}
\vspace{0.15cm}

\textcolor{black}{We follow the paradigm of FSL and adopt a prior base set in our FPML task. In this part, we demonstrate that our method has good robustness to the number of the base categories.} The results are summarized in Fig. \ref{fig:5}. \textcolor{black}{On one hand,} these results indicate that the use of more base visual concepts leads to higher accuracy as rich category information is beneficial for building \textcolor{black}{the} representative anchors. However, \textcolor{black}{on the other hand}, our UGDA can still yield the acceptable performance \textcolor{black}{when there only exist the limited base categories}. For example, our UGDA's accuracy only drops by $2.10 \%$ \textcolor{black}{at the view-missing rate $\eta=0.1$} when the number of the base visual concepts $|\bm{\mathrm{C}}_{b}|$ decreases from $80$ to $5$. \textcolor{black}{Also, the corresponding accuracy drop is still only $4.71 \%$ at \textcolor{black}{the view-missing rate} $\eta=0.3$. Even by considering only two base visual concepts}, our method can still alleviate the negative impact \textcolor{black}{brought by view missing, e.g., the use of UGDA can bring $15.92 \%$ accuracy improvement at $|\bm{\mathrm{C}}_{b}|=2$ under the view-missing rate $\eta=0.3$. Last but not least}, our UGDA can yield relatively stable performance when using different categories to build the base set. Each experiment \textcolor{black}{in Fig. \ref{fig:5}} is repeatedly conducted $5$ times. Moreover, in each time, the categories are randomly chosen to build the base set. We can see that the standard deviation (represented as the light blue and the light red areas \textcolor{black}{in Fig. \ref{fig:5}}) is still acceptable.

\vspace{0.3cm}
\noindent\textcolor{black}{$(c).$ \textit{Robustness Analysis in Scenarios that Base Visual Concepts are Less Relevant to Test Categories}}
\vspace{0.15cm}

We also demonstrate that UGDA can be applied to the scenario that the base categories are less relevant to the test classes. We adopt the two extra datasets tieredImagenet \cite{ren2018meta} and Birds-200-2011 \cite{wah2011caltech} in this part. \textcolor{black}{In the following,} we first briefly introduce these datasets, and then give our experimental results and analysis in detail.

\begin{table*}
    \centering
    \caption{\textcolor{black}{The robustness analysis in the domain-shift setting ``tieredImagenet$\to$Birds-200-2011''. The experiments are conducted on the $5$-way $1$-shot $3$-view setting. Notice that ``SE'' represents the standard error and ``$\eta$'' denotes the view-missing rate.}}
    \setlength{\tabcolsep}{4mm}{
    \begin{tabular}{c|cccccccc}
        \toprule
        \diagbox{Method}{Acc $\pm$ SE (\%)}{$\eta$} & $0.1$ & $0.2$ & $0.3$ & $0.4$ & $0.5$ & $0.6$ \\
                \midrule
        PTMAP \cite{hu2021leveraging} & \cellcolor{gray!15}{$74.55\pm 0.26$}  & \cellcolor{gray!15}{$70.71\pm 0.33$} & $65.66\pm 0.27$ & $59.90\pm 0.16$  & $52.68\pm 0.17$  & $44.08\pm 0.22$\\
        RFS \cite{tian2020rethinking}  & $62.53\pm 0.28$  & $54.30\pm 0.45$ & $47.61\pm 0.39$ & $41.99\pm 0.20$  & $37.75\pm 0.34$  & $35.37\pm 0.13$ \\
        ICMSC \cite{wang2020icmsc} & $58.79\pm 0.34$  & $46.75\pm 0.40$ & $38.71\pm 0.45$ & $32.62\pm 0.23$  & $26.58\pm 0.16$  & $23.05\pm 0.11$ \\
        COMPLETER \cite{lin2021completer} & $68.56\pm 0.33$  & $63.46\pm 0.38$ & $57.63\pm 0.47$ & $51.89\pm 0.21$  & $46.16\pm 0.28$  & $40.30\pm 0.29$ \\
        \midrule
        \midrule
        PTMAP+UGDA & \cellcolor{gray!30}{$75.74\pm 0.35$} & \cellcolor{gray!30}{$73.12\pm 0.42$} & \cellcolor{gray!30}{$70.08\pm 0.40$} & \cellcolor{gray!30}{$67.02\pm 0.42$}  & \cellcolor{gray!15}{$63.61\pm 0.25$}  & \cellcolor{gray!60}{$60.92\pm 0.38$} \\
        RFS+UGDA & $71.70\pm 0.30$  & $69.23\pm 0.40$ & \cellcolor{gray!15}{$66.45\pm 0.42$} & \cellcolor{gray!15}{$63.74\pm 0.42$}  & \cellcolor{gray!30}{$60.73\pm 0.31$}  & \cellcolor{gray!15}{$57.98\pm 0.42$} \\
        UGDA & \cellcolor{gray!60}{$76.72\pm 0.30$} & \cellcolor{gray!60}{$74.03\pm 0.37$} & \cellcolor{gray!60}{$70.78\pm 0.37$} & \cellcolor{gray!60}{$67.42\pm 0.33$} & \cellcolor{gray!60}{$63.90\pm 0.24$} & \cellcolor{gray!30}{$60.42\pm 0.35$} \\
        \bottomrule
    \end{tabular}}
    \label{tab:8}
\end{table*}

\begin{table*}
    \centering
    \caption{\textcolor{black}{The ablation study for the anchor distribution self-rectification. The experiments conducted on Scene15 are in the $3$-way $1$-shot $2$-view setting, and the experiments on Caltech102 are in the $5$-way $1$-shot $6$-view setting. ``$\mathcal{L}_{ce}$'' denotes the cross-entropy term and ``$\mathcal{L}_{se}$'' represents the Shannon-entropy term. The best results of each dataset in the column are highlighted by the gray background \textcolor{gray!60}{$\blacksquare$}.}}
    \setlength{\tabcolsep}{4mm}{
    \begin{tabular}{c|cccccccc}
        \toprule
        \diagbox{Method}{Acc $\pm$ SE (\%)}{$\eta$} & $0$ & $0.1$ & $0.2$ & $0.3$ & $0.4$ & $0.5$ \\
        \midrule
        & \multicolumn{6}{c}{\textcolor{black}{Caltech102 Dataset}} \\
        Full ADSR & \cellcolor{gray!60}{$59.13\pm 0.59$} & \cellcolor{gray!60}{$54.15\pm 0.60$} & \cellcolor{gray!60}{$51.35\pm 0.57$} & \cellcolor{gray!60}{$47.81\pm 0.56$} & \cellcolor{gray!60}{$44.29\pm 0.53$} & \cellcolor{gray!60}{$42.05\pm 0.52$} \\
        w/o $\mathcal{L}_{ce}$ & $55.04\pm 0.62$  & $51.75\pm 0.62$ & $49.09\pm 0.59$ & $45.50\pm 0.59$  & $42.30\pm 0.57$ & $40.14\pm 0.55$ \\
        w/o $\mathcal{L}_{se}$ & $55.91\pm 0.55$ & $50.57\pm 0.58$ & $45.72\pm 0.56$ & $42.07\pm 0.52$ & $39.37\pm 0.48$ & $36.74\pm 0.43$ \\
        w/o $\mathcal{L}_{ce}$ + w/o $\mathcal{L}_{se}$ & $52.86\pm 0.55$  & $49.18\pm 0.54$ & $45.49\pm 0.52$ & $41.85\pm 0.49$ & $39.60\pm 0.47$ & $36.89\pm 0.43$ \\
        \midrule
        \midrule
        & \multicolumn{6}{c}{\textcolor{black}{Scene15 Dataset}} \\
        Full ADSR & \cellcolor{gray!60}{$72.37\pm 0.65$} & \cellcolor{gray!60}{$70.15\pm 0.63$} & \cellcolor{gray!60}{$67.11\pm 0.66$} & \cellcolor{gray!60}{$65.83\pm 0.66$} & \cellcolor{gray!60}{$63.26\pm 0.67$} & \cellcolor{gray!60}{$62.17\pm 0.69$} \\
        w/o  $\mathcal{L}_{ce}$ & $61.90\pm 0.82$  & $61.69\pm 0.75$ & $59.41\pm 0.77$ & $59.11\pm 0.76$  & $56.80\pm 0.77$ & $56.63\pm 0.76$ \\
        w/o  $\mathcal{L}_{se}$ & $69.88\pm 0.63$ & $68.26\pm 0.61$ & $65.53\pm 0.63$ & $63.88\pm 0.63$ & $61.41\pm 0.65$ & $60.46\pm 0.66$ \\
        w/o $\mathcal{L}_{ce}$ + w/o $\mathcal{L}_{se}$ & $65.62\pm 0.62$  & $64.34\pm 0.61$ & $62.45\pm 0.63$ & $61.15\pm 0.64$  & $59.13\pm 0.65$ & $57.97\pm 0.66$ \\
        \bottomrule
    \end{tabular}}
    \label{tab:9}
\end{table*}

\textbf{\textcolor{black}{tieredImagenet and Birds-200-2011}}. tieredImagenet consists of about $700$k images collected from the $608$ categories. Following \cite{tian2020rethinking,he2020memory}, we divide this dataset according to the high-level classes defined in Imagenet \cite{russakovsky2015imagenet}, \textcolor{black}{in which the $351/97/160$ classes are used for building the base set$/$validation set$/$test set}. This high-level split simulates a more realistic few-shot scenario that the categories of the test set and the base set are distinct sufficiently \cite{ren2018meta,he2020memory,tian2020rethinking}. The Birds-200-2011 dataset, \textcolor{black}{collected from the $200$ fine-grained bird species}, contains the $11788$ images. Following the standard division in FSL, this dataset is split into the $100/50/50$ categories for building the base set$/$validation set$/$test set. The Birds-200-2011 dataset has been frequently used in FSL to evaluate the model performance in the domain-shift setting \cite{boudiaf2020information,hu2021leveraging}. \textcolor{black}{For both of these datasets, the deep visual features extracted by MobileNet \cite{howard2017mobilenets}, ResNet-18 \cite{he2016deep}, and WiderRes-28 \cite{zagoruyko2016wide} are used as the three views for each sample.}

\begin{figure*}[t!]
\centering
 \includegraphics[height=5.6cm]{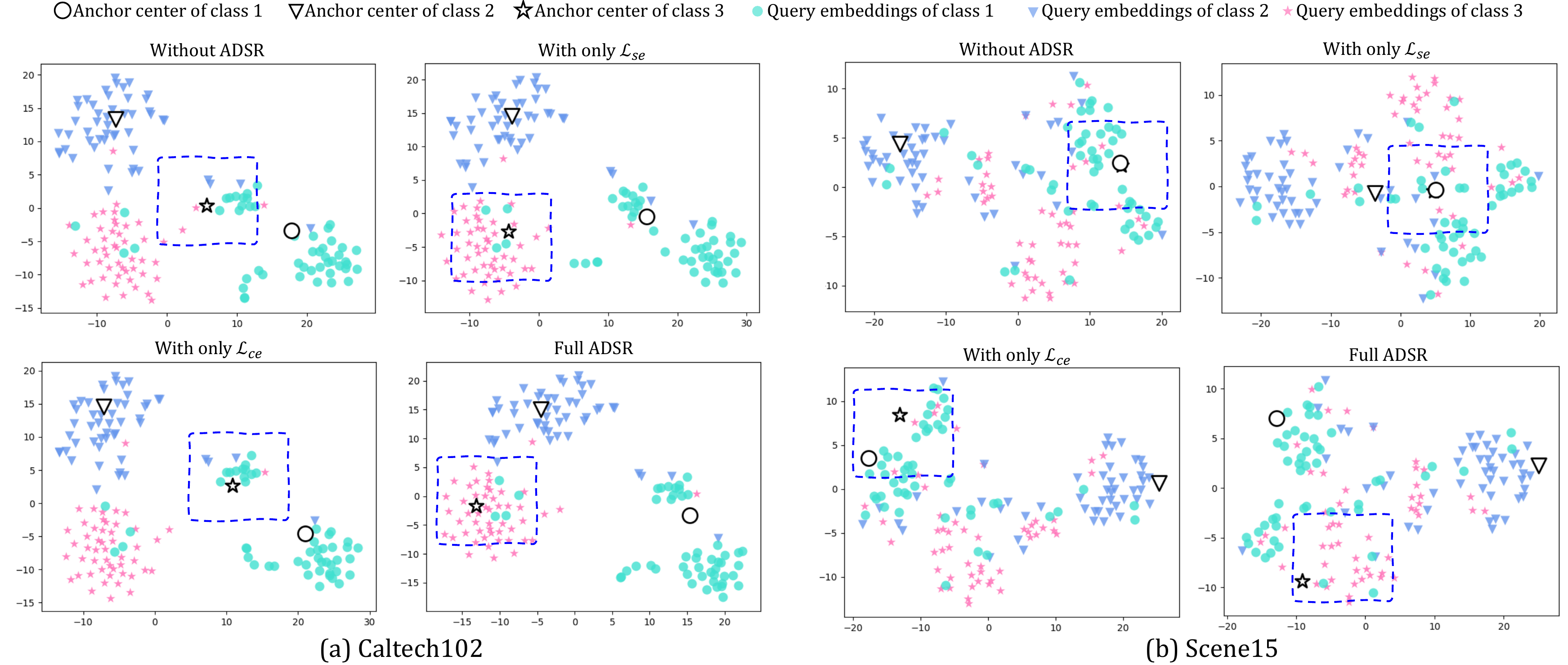}
\caption{\textcolor{black}{The qualitative analysis for our anchor distribution self-rectification. In the figure, the small marks denote the latent embeddings of the queries while the big marks with \textcolor{black}{black} edge lines represent the anchor centers of each category. The visualized analysis on the Caltech102 and the Scene15 dataset is conducted under the $3$-way $1$-shot setting at the view-missing rate $\eta=0.3$}.}
\label{fig:6}
\end{figure*}

\begin{figure*}[t!]
\centering
 \includegraphics[height=3.4cm]{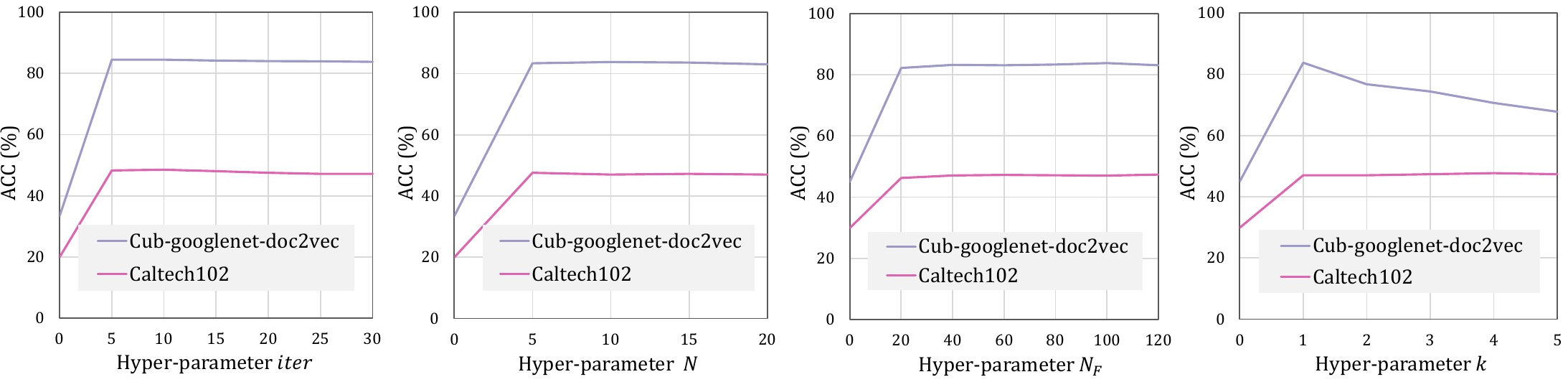}
\caption{\textcolor{black}{The influence of the hyper-parameters $iter$, $N$, $N_{F}$, and $k$. The experiments are conducted on  Caltech102 and Cub-googlenet-doc2vec under the view-missing rate $\eta=0.3$.}}
\label{fig:7}
\end{figure*}

\textbf{\textcolor{black}{Scenario one: base visual concepts are dissimilar to test categories}}. Aiming to demonstrate that our method can be applied to the scenario where the base and the test categories are dissimilar, we experiment UGDA on the tieredImagenet dataset. \textcolor{black}{The base set and the test set of tieredImagenet have  dissimilar categories as they are split according to the high-level categories \cite{ren2018meta,he2020memory,tian2020rethinking}.} From Table. \ref{tab:7}, we can see our method still exhibits high effectiveness. \textcolor{black}{Specifically, our UGDA method shows better robustness to the view-missing issue. For example, at the view-missing rate $\eta=0.6$}, the accuracy of PTMAP, RFS, ICMSC, and COMPLETER is $17.67 \%$, $26.98 \%$, $38.50 \%$, and $21.66 \%$ lower than that of our UGDA. \textcolor{black}{Also, the use of UGDA can consistently boost the robustness of the FSL models to view missing. For example, by equipping PTMAP and RFS with our UGDA, their accuracy can be improved at almost all the view-missing rates listed in the table.} \par

\textbf{Scenario two: there is domain shift between base visual concepts and test categories.} \textcolor{black}{In addition to the low similarity between the base and the test categories, it is often the case that data are from different domains}, which may also have negative effects on the model's effectiveness. \textcolor{black}{Considering this issue}, we also try to apply our method in the domain-shift setting ``tieredImagenet$\to$Birds-200-2011'', which means the base set of tieredImagenet is chosen as the prior knowledge while the test set of Birds-200-2011 is used for testing. The results on Table.~\ref{tab:8} \textcolor{black}{validate the effectiveness of our method as well}. \textcolor{black}{On one hand,} UGDA obviously has better robustness to view missing than that of the compared approaches. \textcolor{black}{On the other hand}, even in the domain-shift setting, UGDA is still able to make the FSL models PTMAP and RFS more robust against the view-missing issue in the low-data scenario.

\subsubsection{\textcolor{black}{Analysis for Our Anchor Distribution Self-rectification}}
\textcolor{black}{In this subsection, we provide the quantitative and the qualitative analysis for the Anchor Distribution Self-Rectification (ADSR), of which the results are summarized in Table.~\ref{tab:9} and Fig.~\ref{fig:6}.} \par

\textbf{Quantitative analysis}. We first conduct the quantitative analysis for ADSR. \textcolor{black}{According to the results in Table.~\ref{tab:9}, we have the following observations. \textcolor{black}{Specifically},} the use of the full ADSR can yield the best performance. Either the removal of the cross-entropy term $\mathcal{L}_{ce}$ (i.e., ``w/o $\mathcal{L}_{ce}$'') or the Shannon-entropy term $\mathcal{L}_{se}$ (i.e., ``w/o $\mathcal{L}_{se}$'') leads to an obvious accuracy drop. \textcolor{black}{This demonstrates the importance of these two terms}. \textcolor{black}{With the supervised cross-entropy term $\mathcal{L}_{ce}$, the unsupervised Shannon-entropy term $\mathcal{L}_{se}$ can constrain the rectification procedure more effectively. Vice versa, under the guidance of $\mathcal{L}_{se}$, $\mathcal{L}_{ce}$ can learn more discriminate anchor representations, which help to build a more effective metric classifier}. \par

\textbf{Qualitative analysis}. \textcolor{black}{The results of the qualitative analysis are shown in Fig.~\ref{fig:6}. In particular,} to clearly reflect the impact of ADSR, the dense anchors are not drawn in the figure. \textcolor{black}{Instead, we visualize the change of the anchor center for each category} (represented as the big marks with black edge lines in Fig.~\ref{fig:6}). \textcolor{black}{The results indicate that ADSR can make the query categorization more accurate}. \textcolor{black}{We take Fig.~\ref{fig:6} (a) as an example. When not using ADSR}, there exists the bias between the anchors of the class $3$ and the queries of the class $1$, \textcolor{black}{which will obviously mislead the classification procedure. By using our ADSR, the anchors become more discriminative to reflect the categories of the queries, e.g., the anchor center of the class $3$ is kept away from the query embeddings of the class $1$, \textcolor{black}{while being closer to the queries that belong to the class $3$}. Obviously, using the cross-entropy term $\mathcal{L}_{ce}$ and the Shannon-entropy term $\mathcal{L}_{se}$ together is better than using  them alone.}

\subsubsection{\textcolor{black}{Hyper-parameter Analysis}}
We also study the influence of the hyper-parameters $iter$, $N$, $k$, and $N_{F}$ for the model. \textcolor{black}{According to the results in Fig.~\ref{fig:7}, we have the following observations. With the increase of $iter$, $N$, or $N_F$, the accuracy of our model improves largely at first. Then, it remains stable and is not obviously affected by these hyper-parameters, indicating the robustness of our method to these hyper-parameters.} Also, we find that the influence of the hyper-parameter $k$ is related to the scale of the base set. On one hand, the base set of Cub-googlenet-doc2vec is small, which only consists of the $6$ base visual categories. Setting $k$ with a large value will reduce the discrimination of the data representations, \textcolor{black}{as the retrieved information for each sample largely comes from the overlapped categories and thereby has poor diversity.} On the other hand, the base set of Caltech102 is larger than that of Cub-googlenet-doc2vec, which contains the $80$ base visual concepts. This facilitates the diversity of the retrieved information. Thus, setting $k$ with a relatively large value does not lower the data's discrimination. Instead, the retrieved rich information can help to build better data representations. Despite of this, we find setting $k$ with a large value does not bring much performance gain, e.g., the accuracy of the setting $k=4$ is only $0.7 \%$ higher than that of $k=1$. Thus, we set $k=1$ on all of the used datasets. The Gaussian anchors are unavailable in the settings $k=0$ and $N_F=0$. \textcolor{black}{Therefore, in these two settings, the classification is conducted by using the original embeddings of the supports and the queries.}\par

\section{Conclusion}
\textcolor{black}{In real-world applications, data scarcity and view missing usually occur simultaneously. Aiming to simulate this challenging scenario}, we propose the few-shot partial multi-view learning task, which aims to overcome the view-missing issue in the low-data regime. To address this task, we propose the unified Gaussian dense-anchoring method, \textcolor{black}{which  anchors the incomplete low-shot data into a unified dense representation space}. In this way, \textcolor{black}{the influence of} data scarcity and view missing can be both alleviated. We hope that our work can attract more attention from the community \textcolor{black}{to this challenge and inspire the proposal of more practical solutions.} We believe the key to this task is to learn the high-quality representations from the incomplete low-shot data. Therefore,  in the future, we plan to advance our anchoring-based approach by fully considering the semantic guidance of the visual concepts.


%

%

\ifCLASSOPTIONcompsoc
  \section*{Acknowledgments}
\else
  \section*{Acknowledgment}
\fi

This work was supported by National Key Research and Development Program under Grant No. 2019YFA0706200, the National Nature Science Foundation of China under Grant No. 62072152, 62172137, 72188101, and the Fundamental Research Funds for the Central Universities under Grant No. PA2020GDKC0023.

\ifCLASSOPTIONcaptionsoff
  \newpage
\fi



%

\bibliographystyle{IEEEtran}
\bibliography{ref}

%

\begin{IEEEbiography}[{\includegraphics[width=1in,height=1.25in,clip,keepaspectratio]{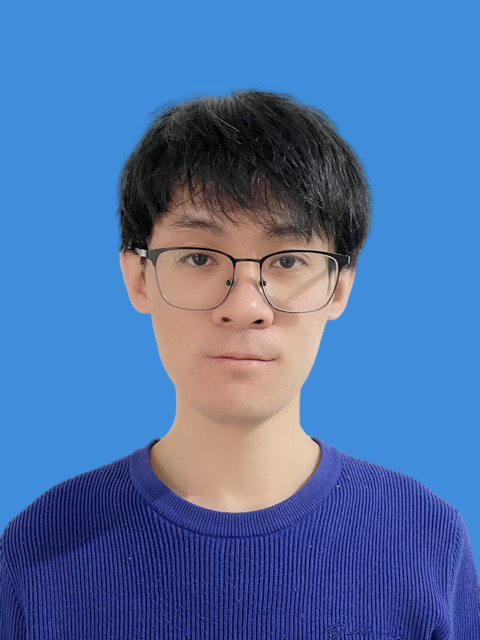}}]{Yuan Zhou}
is pursuing his Ph.D Degree at School of Computer and Information, Hefei University of Technology. He is also with Key Laboratory of Knowledge Engineering with Big Data (Hefei University of technology), Ministry of Education. His research interests include few-shot learning and image segmentation.
\end{IEEEbiography}

\begin{IEEEbiography}[{\includegraphics[width=1in,height=1.25in,clip,keepaspectratio]{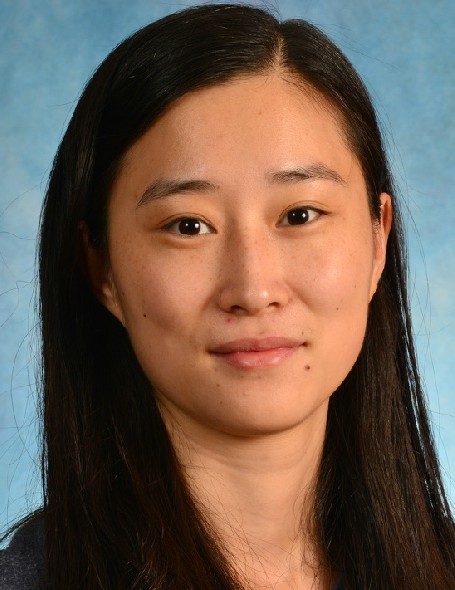}}]{Yanrong Guo}
is a professor at School of Computer and Information, Hefei University of Technology (HFUT). She is also with Key Laboratory of Knowledge Engineering with Big Data (Hefei University of technology), Ministry of Education. She received her Ph.D. degree at HFUT in 2013. She was a postdoc researcher at University of North Carolina at Chapel Hill (UNC) from 2013 to 2016. Her research interests include image analysis and pattern recognition.
\end{IEEEbiography}

\begin{IEEEbiography}[{\includegraphics[width=1in,height=1.25in,clip,keepaspectratio]{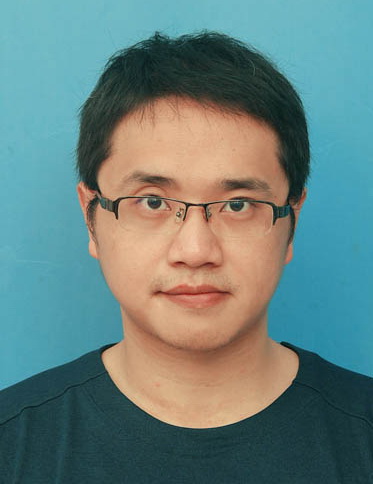}}]{Shijie Hao}
is an associate professor at School of Computer Science and Information Engineering, Hefei University of Technology (HFUT). He is also with Key Laboratory of Knowledge Engineering with Big Data (Hefei University of technology), Ministry of Education. He received his Ph.D. degree at HFUT in 2012. His research interests include image processing and multimedia content analysis.
\end{IEEEbiography}

\begin{IEEEbiography}[{\includegraphics[width=1in,height=1.25in,clip,keepaspectratio]{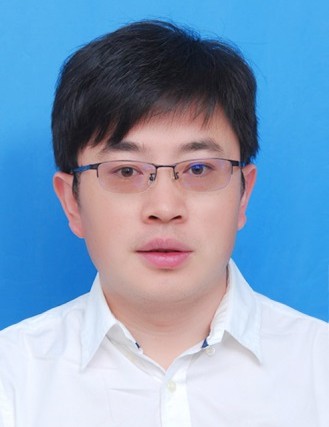}}]{Richang Hong}
received the Ph.D. degree from the University of Science and Technology of China, Hefei, China, in 2008. He was a Research Fellow of the School of Computing with the National University of Singapore, from 2008 to 2010. He is currently a Professor with the Hefei University of Technology, Hefei. He is also with Key Laboratory of Knowledge Engineering with Big Data (Hefei University of technology), Ministry of Education. He has coauthored over 100 publications in the areas of his research interests, which include multimedia content analysis and social media. He is a member of the ACM and the Executive Committee Member of the ACM SIGMM China Chapter. He was a recipient of the Best Paper Award from the ACM Multimedia 2010, the Best Paper Award from the ACM ICMR 2015, and the Honorable Mention of the IEEE Transactions on Multimedia Best Paper Award. He has served as the Technical Program Chair of the MMM 2016, ICIMCS 2017, and PCM 2018. Currently, he is an Associate Editor of IEEE Transactions on Big Data, IEEE Transactions on Computational Social System, ACM Transactions on Multimedia Computing Communications and Applications, Information Sciences (Elsevier), Neural Processing Letter (Springer) and Signal Processing (Elsevier).
\end{IEEEbiography}

\begin{IEEEbiography}[{\includegraphics[width=1in,height=1.25in,clip,keepaspectratio]{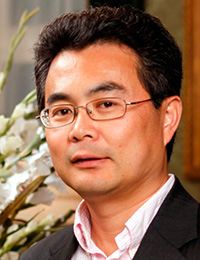}}]{Jiebo Luo}
(S'93-M'96-SM'99-F'09)  joined the Department of Computer Science, University of Rochester, in 2011, after a prolific career of more than 15 years with Kodak Research. He has authored more than 400 technical papers and holds more than 90 U.S. patents. His research
interests include computer vision, machine learning, data mining, social media, and biomedical informatics. He has served as the Program Chair for ACM Multimedia 2010, IEEE CVPR 2012, ACM ICMR 2016, and IEEE ICIP 2017, and on the Editorial Boards of IEEE TPAMI, IEEE TMM, IEEE TCSVT, IEEE TBD, Pattern Recognition, Machine Vision and Applications, and ACM Transactions on Intelligent Systems and Technology. He is also a fellow of IEEE, ACM, AAAI, SPIE, and IAPR.
\end{IEEEbiography}




\end{document}